\documentclass[11pt]{article}
\PassOptionsToPackage{table}{xcolor}
\usepackage{acl}
\usepackage{times}
\usepackage{latexsym}
\usepackage[T1]{fontenc}
\usepackage[utf8]{inputenc}
\usepackage{microtype}
\usepackage{inconsolata}
\usepackage{graphicx}
\usepackage{xcolor}
\usepackage{multirow}
\usepackage{enumitem}
\usepackage{subfigure}
\usepackage{array}
\usepackage{booktabs}
\usepackage{amsmath}
\usepackage{amssymb}
\usepackage{amsthm}
\usepackage{url}

\newtheorem{theorem}{Theorem}
\newtheorem{corollary}[theorem]{Corollary}
\begin{document}

\title{Unveiling Spectral Mechanisms in Training-Free LLM Text Detection}
\author{
  \textbf{Haitong Luo\textsuperscript{1}},
  \textbf{Xuying Meng\textsuperscript{1}\thanks{Corresponding authors.}},
  \textbf{Weiyao Zhang\textsuperscript{1}},
  \textbf{Wenji Zou\textsuperscript{1}},
\\
  \textbf{Shengfeng Lou\textsuperscript{1}},
  \textbf{Xuefeng Jiang\textsuperscript{1}},
  \textbf{Chungang Lin\textsuperscript{1}},
  \textbf{Yujun Zhang\textsuperscript{1}\thanks{Corresponding authors.}}
\\
\\
  \textsuperscript{1}Institute of Computing Technology, Chinese Academy of Sciences
\\
  \small{
    \texttt{luohaitong21s@ict.ac.cn}
  }
}

\maketitle

\begin{abstract}
The rapid advancement of Large Language Models (LLMs) makes it increasingly difficult to distinguish human writing from machine-generated text. Training-free detection offers a scalable solution, yet common confidence-based metrics mainly measure average token probabilities and often miss the signal fluctuations that characterize human writing, which we call ``generative vitality''. Spectral analysis offers a way to capture this vitality, but its mechanism and practical boundaries remain underexplored. In this paper, we analyze spectral detection from both theoretical and empirical perspectives. We connect spectral energy to variance in proxy log-probability trajectories and explain how broader human token choices create the fluctuations used by frequency-domain indicators. We further show that the strength of this signal depends on text length and sampling range: spectral evidence is clearest for long, continuous, constrained generation, while short, fragmented, mixed, and edited settings require complementary confidence and fluctuation views. These findings clarify when frequency-domain detection works and provide guidance for future multi-dimensional detector design.
\end{abstract}
\section{Introduction}

\begin{figure}[t]
    \centering
    \includegraphics[width=0.8\linewidth]{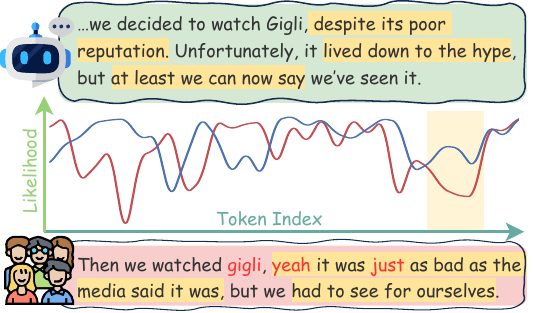}
    \caption{
    \textbf{Illustration of Generative Vitality.} The plot shows token log-likelihood trajectories computed by a proxy LLM. Human writing (red) contains vitality spikes caused by unpredictable low-probability tokens, marked in \textcolor{red}{red} in the text below. AI outputs (blue) are typically smoother because decoding favors high-probability tokens. Yellow-highlighted regions mark segments with large generation-probability divergence.}
    \label{fig:concept}
\end{figure}
The rapid advancement of Large Language Models (LLMs) has transformed digital content creation~\cite{crothers2023machine}, while also blurring the distinction between texts written by humans and those generated by machines. This ambiguity creates risks such as misinformation~\cite{opdahl2023trustworthy,fang2024bias} and academic plagiarism~\cite{else2023chatgpt,currie2023academic}. Among existing solutions, \textbf{training-free detection}~\cite{solaiman2019release,su2023detectllm,gehrmann2019gltr,ippolito2019automatic,xu2024training,luo2025specdetect} is particularly attractive because it identifies LLM-generated text from statistical patterns such as probability curvature~\cite{bao2023fast, mitchell2023detectgpt} or entropy~\cite{ippolito2019automatic}, without requiring large supervised training sets.

Most training-free methods rely on \textbf{confidence-based indicators} such as LogLikelihood and Rank~\cite{solaiman2019release}, which assume that LLMs select tokens with higher probabilities than humans. Recent studies~\cite{xu2024training,luo2025specdetect} suggest that these metrics overlook signal \textbf{fluctuation}, termed \textbf{generative vitality}. We define this vitality as the intermittent emergence of \textbf{unpredictable tokens} from a low-probability region. As visualized in Figure~\ref{fig:concept}, human authors often introduce expressive low-probability tokens that create sharp negative spikes in the likelihood trajectory, while LLM decoding more often follows a smoother high-probability path.

To capture these fluctuations, initial efforts such as Lastde~\cite{xu2024training} use multi-scale entropy, but this can introduce sensitivity to hyperparameters. SpecDetect~\cite{luo2025specdetect} takes a frequency-domain approach, based on the observation that human writing tends to exhibit higher \textbf{spectral energy} than LLM-generated text. Although SpecDetect performs well, the mechanism behind this spectral gap and its behavior in complex real-world scenarios, such as short snippets, mixed-source content~\cite{lee2022coauthor,wang2024semeval}, and collaborative editing~\cite{zhang2024llm}, remain insufficiently understood. Consequently, the practical utility of frequency-domain detection is constrained by three critical ``unknowns'' that we address in this paper.

\textbf{The Theoretical Unknown:} \textit{Why are frequency-domain methods effective?} Empirical results show a spectral energy gap, but the generation source of this gap remains unclear. \textbf{The Capability Unknown:} \textit{What are the boundaries of their effectiveness?} Existing evaluations are primarily document-level, leaving short text, mixed-source text, and human-AI editing underexplored. \textbf{The Solution Unknown:} \textit{How should we address spectral failure mode?} If spectral methods cannot excel in all scenarios, what is the solution? Specifically, can we integrate them with traditional methods to build a robust system?

In this work, we present a comprehensive analysis to bridge these gaps. First, we develop a theoretical model centered on the generative vitality of human writing and show how fluctuation disparities manifest as a frequency-domain signature. Building on this foundation, we evaluate spectral evidence across standard document-level benchmarks and more challenging real-world scenarios, including mixed-source and collaborative writing.

Our analysis leads to several key insights. \textbf{First}, spectral and confidence-based metrics are mathematically and empirically distinct, reflecting fluctuation and mean-level properties of the proxy probability signal. \textbf{Second}, spectral evidence is strongest when generation is long, continuous, and produced under a constrained sampling scope, while short fragments and point-wise editing weaken the signal. \textbf{Finally}, integrating confidence metrics is a potential direction for scenarios where frequency-domain methods fail, while naive combination can dilute useful signals. Developing more effective, adaptive fusion strategies remains a critical challenge for future research.

By mapping out where frequency-domain detection works and which signals become informative in different regimes, this paper provides a roadmap for future multi-dimensional AI detectors. Our code is available in this \href{https://anonymous.4open.science/r/Unveiling-Spectral-Mechanisms-in-Training-Free-LLM-Text-Detection-F841}{repository}.

\section{Related Work}
\label{sec:related_work}
We provide a concise overview here and include a detailed discussion in Appendix~\ref{app:extended_related_work}.

\noindent\textbf{Training-Free LLM Text Detection.}
Training-free detection identifies machine-generated text through intrinsic statistical signatures. Existing methods include distribution-based approaches that compare an input with perturbed or generated variants~\cite{mitchell2023detectgpt,su2023detectllm,yang2023dna,bao2023fast}, and sample-based methods that directly score token-level statistics~\cite{solaiman2019release,su2023detectllm,gehrmann2019gltr,ippolito2019automatic,xu2024training,luo2025specdetect}. Since sample-based indicators expose the underlying probability signal most directly, our study follows this branch and analyzes how frequency-domain detection models probability-signal fluctuation, with SpecDetect~\cite{luo2025specdetect} serving as a representative spectral method.

\noindent\textbf{Detection in Complex Scenarios.}
Real-world content often contains mixed-source text~\cite{lee2022coauthor,wang2024semeval} and collaborative authorship~\cite{zhang2024llm}, where human and machine segments are interleaved or edited together. Recent studies address these settings with supervised sequence-labeling or boundary-detection models~\cite{zeng2024detecting,su2025haco,kadiyala2025robust}, while training-free detectors are still mainly evaluated on long-form, single-source documents. We examine the utility and limitations of frequency-domain indicators when signals are fragmented by mixing or reshaped by collaborative editing.

\section{Detection Signals and Spectral Mechanism}
\label{sec:pipeline}
\label{sec:theory}

\begin{figure}[t]
    \centering
    \includegraphics[width=0.9\linewidth]{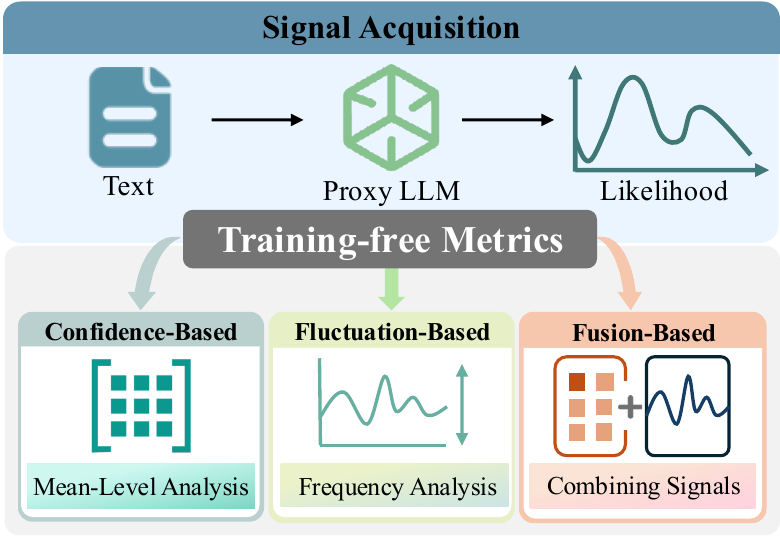}
    \caption{Overview of the Training-Free Detection Pipeline.}
    \label{fig:framework}
\end{figure}

In the training-free LLM detection paradigm, text classification is transformed into mining patterns within the token probability sequence. As illustrated in Figure~\ref{fig:framework}, a proxy LLM first converts text into token log-probabilities, and detectors extract features from this signal.

\subsection{From Text to Probability Signals}
\label{sec:signal_gen}

Let $X = \{w_1, w_2, \dots, w_N\}$ be a sequence of $N$ tokens. To analyze the generation traces, existing works utilize a proxy LLM $\mathcal{M}$ (e.g., GPT-J~\cite{gpt-j}) to compute the step-wise conditional probabilities. We define the probability signal as the log-probability of the token:
\begin{equation}
    x[t] = \log P_{\mathcal{M}}(w_t \mid w_{<t})
\end{equation}
which forms the probability signal analyzed by training-free detectors. Given this signal, a detection algorithm computes an indicator $y=f(x)$, where larger $y$ indicates stronger evidence of machine origin. An effective detector ranks machine text above human text in expectation, i.e., $\mathbb{E}[f(x_{\text{LLM}})] > \mathbb{E}[f(x_{\text{Human}})]$.

\subsection{Metric Families: Confidence and Fluctuation}
\label{sec:taxonomy}

To evaluate how different methods extract information from $x[t]$, we categorize prevalent training-free metrics into a unified taxonomy in Table~\ref{tab:metrics_taxonomy}. We distinguish confidence-oriented indicators from fluctuation-oriented indicators; detailed formulations appear in Appendix~\ref{app:metric_details}, and empirical relationships are examined in Section~\ref{sec:orthogonality}.

\textbf{Confidence-based Metrics.} Most existing methods focus on the global elevation of log-probabilities. Indicators such as LogLikelihood, LogRank~\cite{solaiman2019release}, LRR~\cite{su2023detectllm}, and Entropy~\cite{gehrmann2019gltr} capture the mean-shift signature, but are often insensitive to the structural rhythm of the text.

\textbf{Fluctuation-based Metrics.} Frequency-domain methods such as SpecDetect~\cite{luo2025specdetect} measure signal fluctuation. As derived below, LLM decoding suppresses human-like surprisal spikes, creating a detectable spectral signature.

\textbf{Fusion-based Metrics.} Hybrid methods combine confidence and fluctuation cues. Lastde~\cite{xu2024training} combines likelihood with multi-scale entropy. We include SpecFusion as a probing indicator: it standardizes and sums SpecDetect and LogLikelihood to test whether mean-level and fluctuation evidence are complementary.

\begin{table}[h]
\centering
\caption{\textbf{Taxonomy of training-free metrics.} We classify metrics by their statistical focus using checkmarks ($\checkmark$) and cross marks ($\times$). \textbf{Conf.} indicates confidence-based metrics, while \textbf{Fluct.} indicates fluctuation-based metrics.}
\label{tab:metrics_taxonomy}
\resizebox{\linewidth}{!}{%
\begin{tabular}{l l c c}
\toprule
\textbf{Metric} & \textbf{Statistical Interpretation} & \textbf{Conf.} & \textbf{Fluct.} \\
\midrule
LogLikelihood (LL) & Measures average token log-probability. & $\checkmark$ & $\times$ \\
LogRank (LR) & Average of log-rank scores for token positions. & $\checkmark$ & $\times$ \\
LRR & Ratio of log-likelihood to log-rank (relative confidence). & $\checkmark$ & $\times$ \\
Entropy & Average predictive uncertainty of the distribution. & $\checkmark$ & $\times$ \\
\midrule
SpecDetect & Frequency-domain modeling of fluctuations. & $\times$ & $\checkmark$ \\
\midrule
Lastde & Ratio of log-likelihood to multi-scale entropy. & $\checkmark$ & $\checkmark$ \\
SpecFusion & Combination of LogLikelihood and SpecDetect. & $\checkmark$ & $\checkmark$ \\
\bottomrule
\end{tabular}%
}
\end{table}

\subsection{Modeling Generative Vitality}
\label{sec:modeling}



We define generative vitality~\cite{luo2025specdetect} as intermittent visits to the scoring model's low-probability tail. Let $x[t]=\log P_{\mathcal{M}}(w_t\mid w_{<t})$ be the log-probability assigned by scoring model $\mathcal{M}$ at position $t$. At each position, $\mathcal{M}$ induces a context-dependent distribution over vocabulary $\mathcal{V}$. For a Top-$p$ threshold $\tau$, the \textbf{Head Set} $\mathcal{V}_{\mathrm{head}}^{(t)}(\tau)$ is the smallest probability-sorted prefix whose cumulative mass reaches $\tau$, while the \textbf{Tail Set} $\mathcal{V}_{\mathrm{tail}}^{(t)}(\tau)$ contains the remaining lower-probability, high-surprisal tokens. Top-$k$ and related truncated decoding rules induce the same type of partition: they differ only in how the cutoff is selected, while our analysis uses the resulting high-probability region versus its complement. This local partition defines a signal-space boundary $\mathcal{L}_{\mathrm{bound}}^{(t)}=\min_{w\in \mathcal{V}_{\mathrm{head}}^{(t)}(\tau)} \log P_{\mathcal{M}}(w\mid w_{<t})$.

\noindent\textbf{A Unified Mixture View.}
The head/tail event can be written with a single source-level mixture. Let $s\in\{H,\mathrm{AI}\}$ denote the text source. For each token, define
\begin{equation}
    x_s[t]
    =
    (1-b_t^{(s)})x_{\mathrm{head}}^{(s)}[t]
    +
    b_t^{(s)}x_{\mathrm{tail}}^{(s)}[t],
\end{equation}
where $b_t^{(s)}=\mathbb{I}\{w_t\in\mathcal{V}_{\mathrm{tail}}^{(t)}(\tau)\}$ indicates whether the token falls in the scoring model's tail set. The tail-token rate is $\gamma_s=\frac{1}{N}\sum_{t=1}^{N}b_t^{(s)}$, the fraction of positions where text source $s$ enters the tail set.

Human and LLM text differ in this tail-token rate: human authors may choose locally unexpected tokens for specificity or style, whereas LLM decoding favors high-probability continuations. In a white-box setting where the scoring and source models are the same, truncated decoding keeps generated tokens mostly inside the head set; when the proxy and source models differ, some generated tokens may enter the proxy-defined tail. Since LLMs still tend toward high-probability tokens, sample-based detection methods commonly rely on the condition $\gamma_H>\gamma_{\mathrm{AI}}$.

\begin{figure}[t]
    \centering
    \includegraphics[width=\linewidth]{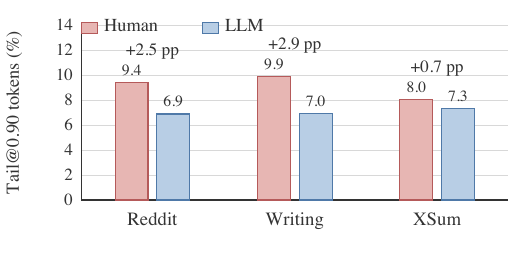}
    \caption{\textbf{Preliminary head/tail check.} Bars show the Tail@0.90 rate under GPT-J-6B proxy scoring. Human continuations generally use more tokens outside the Top-$p=0.90$ head window than paired LLM continuations, especially on WritingPrompts and Reddit. The gap is smaller on XSum.}
    \label{fig:headtail_prelim}
\end{figure}

We examine this condition on the three standard datasets in Section~\ref{sec:setup}, using GPT-J-6B and Llama3-8B as proxy scoring models. We measure two tail-entry rates: the fraction of tokens outside the Top-$p=0.90$ head set and the fraction whose proxy rank is greater than 50. Figure~\ref{fig:headtail_prelim} shows the GPT-J-6B Tail@0.90 result, and Appendix~\ref{app:headtail} reports GPT-J-6B Rank$>$50 and the Llama3-8B counterparts in Figures~\ref{fig:headtail_gptj_rank50}, \ref{fig:headtail_llama3_tail90}, and \ref{fig:headtail_llama3_rank50}. Even under black-box proxy--source mismatch, human text enters the proxy tail region more often than AI text.

\noindent\textbf{The Fluctuation Divergence.}
\label{sec:derivation}
This modeling enables a formal comparison of volatility across human and LLM generation processes.

\begin{theorem}[The Fluctuation Divergence]
\label{thm:variance}
Let $\sigma_H^2=\operatorname{Var}(x_H[t])$ and $\sigma_{\mathrm{AI}}^2=\operatorname{Var}(x_{\mathrm{AI}}[t])$. The human--AI fluctuation divergence is $\Delta\sigma^2=\sigma_H^2-\sigma_{\mathrm{AI}}^2$. Under the unified mixture view, with shared head/tail conditional statistics across text sources,
\begingroup
\small
\setlength{\abovedisplayskip}{1pt}
\setlength{\belowdisplayskip}{1pt}
\setlength{\abovedisplayshortskip}{0pt}
\setlength{\belowdisplayshortskip}{1pt}
\begin{align}
    \Delta\sigma^2
    &=
    (\gamma_H-\gamma_{\mathrm{AI}})
    (\sigma_{\mathrm{tail}}^2-\sigma_{\mathrm{head}}^2) \nonumber\\
    &\quad+
    \bigl[\gamma_H(1-\gamma_H) \nonumber\\
    &\qquad
    -\gamma_{\mathrm{AI}}(1-\gamma_{\mathrm{AI}})\bigr]
    (\mu_{\mathrm{tail}}-\mu_{\mathrm{head}})^2 .
\end{align}
\endgroup
\end{theorem}

Proof details are in Appendix~\ref{app:proof_variance}. The formula explains why human writing is expected to have larger fluctuation under the tail-rate condition above. The first term is positive when human text enters the tail more often and tail-region log-probabilities are more dispersed than head-region values. The second term comes from the mean separation between the two regions: head tokens have log-probabilities close to zero, whereas tail tokens have much lower values. In the usual sparse-tail regime, $\gamma_H>\gamma_{\mathrm{AI}}$ makes this mean-separation term positive as well, so both terms increase $\sigma_H^2$ relative to $\sigma_{\mathrm{AI}}^2$.

\subsection{Spectral Energy as a Fluctuation Signature}
\label{sec:spectral_proof}

The variance gap in Theorem~\ref{thm:variance} transfers to spectral energy through Parseval's identity.


\begin{corollary}[Spectral Energy Gap]
\label{thm:energy}
For a centered signal $z_s[t]=x_s[t]-\mu_s$, let $\bar{\mathcal{E}}_s=\frac{1}{N}\sum_{k=0}^{N-1}|\mathcal{F}(z_s)_k|^2$ be its average spectral energy. Under the same mixture view,
\begin{align}
    \Delta\bar{\mathcal{E}}
    &=
    \mathbb{E}[\bar{\mathcal{E}}_H]
    -
    \mathbb{E}[\bar{\mathcal{E}}_{\mathrm{AI}}]
    =
    N\Delta\sigma^2 .
\end{align}
\end{corollary}

The proof is in Appendix~\ref{app:proof_energy}. For the centered probability signal, Parseval's identity~\cite{hardy1931note} maps the larger time-domain variance of human text to larger expected spectral energy. LLM decoding suppresses these fluctuations, yielding lower expected spectral energy for machine-generated text. SpecDetect therefore uses negative energy as the AI-likeness score so larger values indicate stronger machine evidence.

\textbf{Impact of Sequence Length.}
Sequence length $N$ controls how reliably spectral evidence can be observed. Short texts may contain too few tail events or vitality spikes, while longer texts give the variance gap more opportunities to appear in the probability trajectory. This motivates the length sensitivity study in RQ2.

\textbf{Impact of Sampling Scope.}
Sampling scope affects $\Delta\sigma^2=\sigma_H^2-\sigma_{\mathrm{AI}}^2$. Restricted decoding keeps machine text in high-probability regions, whereas larger Top-$p$, Top-$k$, or temperature lets generated text enter the tail more often, raising $\sigma_{\mathrm{AI}}^2$ and narrowing the spectral gap. This motivates the sampling-scope analysis in RQ3.

\section{Empirical Evaluation in Standard Scenarios}
\label{sec:baseline}

Having established the spectral mechanism, we now transition to empirical verification. This section evaluates whether frequency-domain indicators capture a distinct analytical dimension and tests the boundary conditions governing their effectiveness, especially sequence length and sampling scope. Specifically, \textbf{RQ 1} asks what relationship frequency-domain analysis has with existing methods and whether it captures a \textbf{distinct analytical dimension} as hypothesized; \textbf{RQ 2} asks how \textbf{sequence length} affects spectral methods compared with other indicators, including the strengths and vulnerabilities of short versus long texts; and \textbf{RQ 3} asks how spectral methods perform when the \textbf{fluctuation scope} of machine text is expanded through stochastic sampling, such as high Temperature, Top-$k$, or Top-$p$.

\subsection{Experimental Setup}
\label{sec:setup}

Our setup follows established document-level zero-shot protocols~\cite{xu2024training,luo2025specdetect}. We evaluate three English datasets: \textbf{XSum}~\cite{narayan2018don} for BBC news summarization, \textbf{WritingPrompts}~\cite{fan2018hierarchical} for creative story generation, and \textbf{Reddit ELI5}~\cite{fan2019eli5} for explanatory question answering; WMT16~\cite{bojar2016findings} is in Appendix~\ref{app:crosslingual_domain} for language transfer checks. Each English dataset contains 150 human-written examples. We use the first 30 tokens of each human text as a prompt and generate a paired machine continuation, yielding balanced 150/150 human-machine evaluation sets with shared prompts and matched lengths.

Following previous work~\cite{xu2024training,luo2025specdetect}, we evaluate all methods under a realistic black-box setting where the detector has no access to the source-model internals or identity. The main experiments use Llama2-13B~\cite{touvron2023llama2} as the source model and GPT-J-6B~\cite{gpt-j} as the proxy model for extracting $x[t]$. Additional source-model details are provided in Appendix~\ref{app:real_protocol}. Additional proxy-model results are reported in Appendix~\ref{app:llama3_proxy}. Unless specified otherwise, generation uses Temperature $T=1.0$, Top-$p=1.0$, Top-$k=50$, and 150-word continuations. Following prior work~\cite{xu2024training,luo2025specdetect}, we rank human and machine samples by detector score and report Area Under the ROC Curve (AUC)~\cite{faraggi2002estimation}.

\subsection{RQ 1: Validation of Metric Distinctness}
\label{sec:orthogonality}

For RQ1, we compute Spearman correlations~\cite{wissler1905spearman} and Principal Component Analysis (PCA)~\cite{abdi2010principal} over detector scores. Figure~\ref{fig:metric_analysis} (Metric Relations on Writing Dataset) shows Writing results; other datasets appear in Appendix~\ref{app:metric_validation}, and input-granularity preferences are discussed in Appendix~\ref{app:ablation_proxy}.

\begin{figure}[t]
    \centering
    \includegraphics[width=1.0\linewidth]{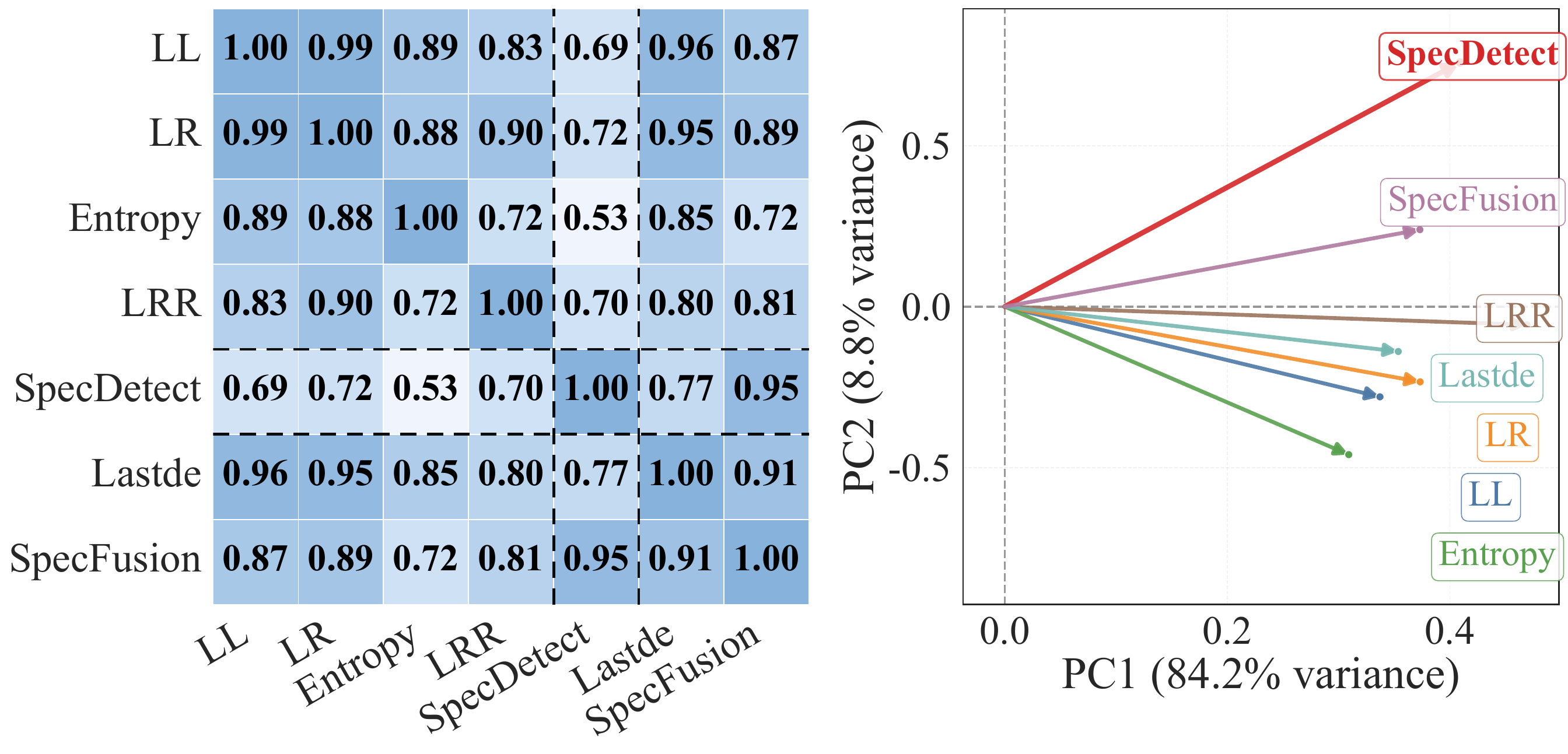}
    \caption{\textbf{Metric Relations on Writing Dataset.} (Left) \textbf{Correlation Heatmap}: Metrics cluster into Confidence and Fluctuation families. (Right) \textbf{PCA Loading Plot}: PC1 (84.2\%) and PC2 (8.8\%) capture the dominant variance directions.}
    \label{fig:metric_analysis}
\end{figure}

\paragraph{Distinct Detection Dimension} 
Figure~\ref{fig:metric_analysis} shows separation between confidence and fluctuation indicators. In the heatmap, LogLikelihood, LogRank, and Entropy cluster together, whereas SpecDetect has lower correlation with this confidence group. The PCA view confirms this: PC1 (84.2\%) captures the common detection signal, while PC2 (8.8\%) separates SpecDetect from mean-probability metrics. This supports the claim that spectral methods capture a fluctuation signature distinct from average probability level.

\subsection{RQ2: Sensitivity to Sequence Length}
\label{sec:exp_length}

\begin{figure}[t]
    \centering
    \includegraphics[width=\linewidth]{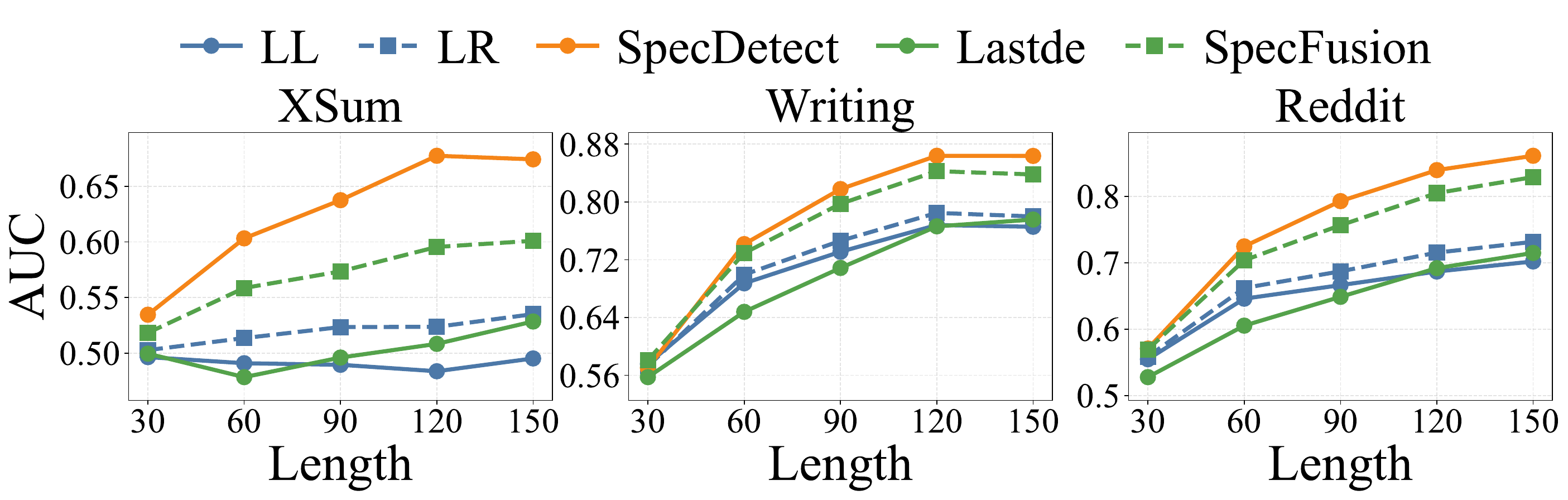}
    \caption{\textbf{Performance vs. Sequence Length on Three Datasets.} Fluctuation-based metrics achieve more significant gains as sequence length increases.}
    \label{fig:length_impact}
\end{figure}

Guided by Section~\ref{sec:theory}, RQ2 truncates test samples to $L \in \{30, 60, 90, 120, 150\}$ words and tracks detector AUC in Figure~\ref{fig:length_impact}. The figure highlights representative metrics. Complete main-setting length curves are reported in Appendix~\ref{app:sensitivity_length}. Additional source-model length results are reported in Appendix~\ref{app:source_length}. The Llama3-8B proxy counterpart is reported in Appendix~\ref{app:llama3_length}. The cross-language length check is reported in Appendix~\ref{app:crosslingual_domain}.

\paragraph{Length Dependent Scaling} 
Figure~\ref{fig:length_impact} shows that spectral detection benefits from longer context. At $L=30$, SpecDetect is weak because short samples may contain too few tail events for a stable fluctuation estimate. As length grows, these local probability changes become easier to observe, and fluctuation-based indicators gain more than confidence-based metrics; fusion indicators typically fall between the two families.

\subsection{RQ3: Sensitivity to Sampling Scope}
\label{sec:exp_decoding}

\begin{figure}[t]
    \centering
    \includegraphics[width=\linewidth]{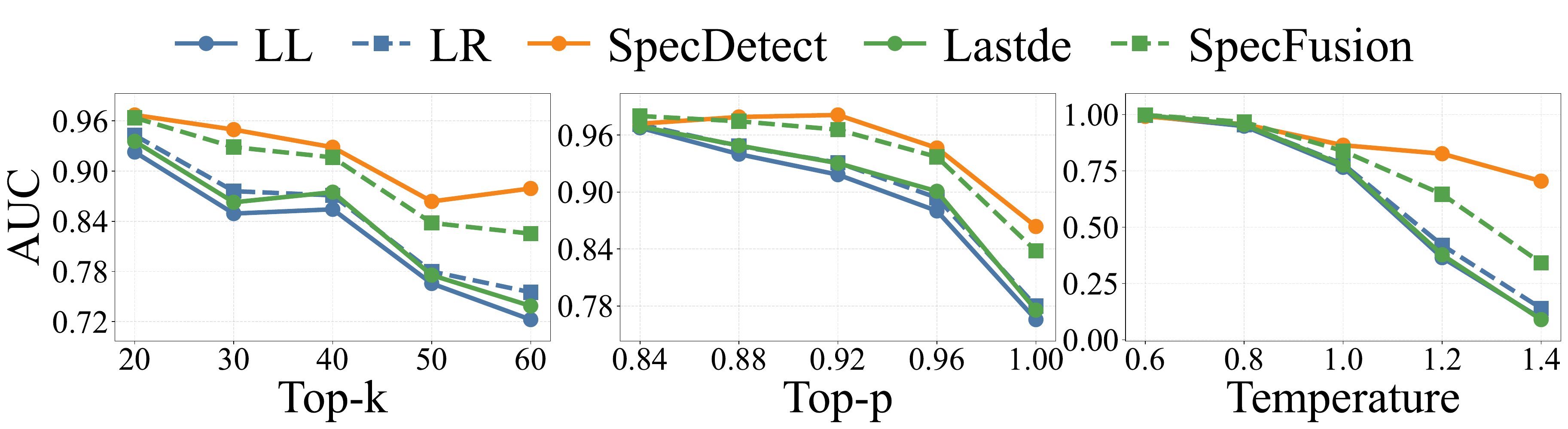}
    \caption{\textbf{Detection Robustness vs. Sampling Parameters on XSum.} Performance decays across all methods as Top-$k$, Top-$p$, and Temperature increase, yet fluctuation-based indicators demonstrate superior resilience compared to confidence-based metrics.}
    \label{fig:full_sensitivity}
\end{figure}

For RQ3, we vary stochastic decoding parameters around the default setting (Top-$k=50$, Top-$p=1.0$, Temperature $T=1.0$): Top-$k \in \{20, \dots, 60\}$, Top-$p \in \{0.84, \dots, 1.0\}$, and $T \in \{0.6, \dots, 1.4\}$. Figure~\ref{fig:full_sensitivity} highlights representative metrics. Complete main-setting sampling curves are reported in Appendix~\ref{app:sensitivity_sampling}. Additional source-model sampling results are reported in Appendix~\ref{app:source_sampling}. The Llama3-8B proxy sampling version is reported in Appendix~\ref{app:llama3_sampling}. The cross-language decoding check is reported in Appendix~\ref{app:crosslingual_domain}.


\paragraph{Stochastic Decoding as a Bottleneck} 
Figure~\ref{fig:full_sensitivity} shows a decline as sampling becomes broader. Increasing $k$, $p$, or $T$ lets generated text access the tail region more often, narrowing the human--machine fluctuation gap and reducing AUC across methods. At extreme temperatures ($T>1.2$), machine text can become more volatile than human text, producing performance inversion; this conditional observation is consistent with the variance-energy mechanism in Corollary~\ref{thm:energy}.


\paragraph{Spectral Resilience vs. Confidence Degradation} 
All indicators suffer under high entropy, yet fluctuation-based metrics degrade more slowly than confidence-based metrics in this setting. As sampling randomness increases, LogLikelihood becomes less distinguishable from human writing, while spectral signatures retain part of the fine-grained variance signal. Fusion indicators generally sit between the two families because they aggregate confidence and fluctuation views.

\subsection{Concluding Takeaways}
\label{sec:standard_takeaways}

The standard-scenario evaluation yields three takeaways. \textbf{Distinct Statistical Properties}: frequency-domain indicators capture fluctuation signatures that are distinct from confidence-based mean-probability metrics. \textbf{Boundary Conditions}: spectral detection benefits from sufficient sequence length, but degrades as stochastic decoding increases machine variance toward, or beyond, human levels. \textbf{Metric Complementarity}: confidence and fluctuation metrics provide safety nets across regimes; spectral structure remains useful when confidence signals flatten under high-entropy decoding, while confidence indicators help when short texts provide too little context for stable fluctuation evidence.

\section{Empirical Evaluation in the Wild}
\label{sec:exp_real}

Following the document-level analysis, we evaluate spectral methods in real-world regimes where human and AI-authored sentences can be interleaved~\cite{wang2024semeval,lee2022coauthor}, edited, or humanized~\cite{zhang2024llm}. We ask two questions: \textbf{RQ1 (Mixed Source Text)} studies how structural mixing affects spectral detection when the signal is fragmented, and \textbf{RQ2 (Collaborative Text)} studies how specific editing operations reshape the fluctuations used for detection. Our goal is to identify when frequency-domain evidence remains useful and when confidence-based cues become a necessary complement.

\subsection{Experimental Setup}

\paragraph{Scenarios and Datasets.}
\label{sec:real_data}

Dataset statistics are provided in Appendix~\ref{app:dataset_details}. \textbf{Scenario I: Mixed-Source Text} covers documents where human-authored and AI-generated sentences coexist: \textbf{SemEval}~\cite{wang2024semeval} uses a human prefix followed by a GPT-3 continuation with variable transition points and ratios, while \textbf{CoAuthor}~\cite{lee2022coauthor} interleaves Human, LLM, and Human-LLM Collaborative sentences throughout a document.

\textbf{Scenario II: Collaborative Text} examines how editing operations transform a text's statistical signature. CoAuthor provides a sentence-level view through its Human-LLM Collaborative category, while \textbf{MixText}~\cite{zhang2024llm} provides an operation-level benchmark with 300 original samples per direction. MixText includes \textbf{AI-Polishing} (Human $\to$ AI), covering token/sentence \textit{Polish}, \textit{Rewrite}, and \textit{Complete}; and \textbf{Humanizing} (AI $\to$ Human), covering token/sentence \textit{Humanize} and \textit{Adapt}, generated by GPT-4 and Llama-2-70b.



\paragraph{Implementation and Evaluation.}
\label{sec:real_protocol}

Following the document-level settings, we use GPT-J-6B~\cite{gpt-j} as the proxy model for all metrics. Scenario I reports the average within-document AUC from sentence-level scores, while Scenario II uses MixText pairwise accuracy: the percentage of pairs where the score correctly shifts after editing, such as an AI-polished text receiving a higher machine-origin score than its human original. Additional protocol details are in Appendix~\ref{app:real_protocol}.

\subsection{RQ 1: Performance in Mixed Source Text}
\label{sec:scenario_mixed}


SemEval focuses on Human (H) vs. LLM (L) segments, while CoAuthor adds a Collaborative (C) category and supports H-L, H-C, and L-C comparisons. Table~\ref{tab:mixed_results} reports the results, and the Llama3-8B proxy counterpart is in Appendix~\ref{app:llama3_table3}. Lastde is omitted from this sentence-level table because its multi-scale diversity entropy component requires longer sequences than sentences provide.
\begin{table}[t]
\centering
\caption{\textbf{Sentence-level Detection across Pure and Collaborative Text.} Average within-document AUC for binary classification among Human (H), LLM (L), and Collaborative (C) sentences. Best results are \textbf{bolded}; second-best are \underline{underlined}.}
\label{tab:mixed_results}
\resizebox{0.85\columnwidth}{!}{%
\begin{tabular}{lcccc}
\toprule
\multicolumn{1}{c}{\multirow{2}{*}{\textbf{Metric}}} & \multicolumn{2}{c}{\textbf{Pure Generation}} & \multicolumn{2}{c}{\textbf{Collaborative Mixed}} \\
\cmidrule(lr){2-3} \cmidrule(lr){4-5}
 & \textbf{SemEval} & \multicolumn{1}{c}{\begin{tabular}{@{}c@{}}\textbf{CoAuthor}\\\textbf{(H vs. L)}\end{tabular}} & \multicolumn{1}{c}{\begin{tabular}{@{}c@{}}\textbf{CoAuthor}\\\textbf{(H vs. C)}\end{tabular}} & \multicolumn{1}{c}{\begin{tabular}{@{}c@{}}\textbf{CoAuthor}\\\textbf{(L vs. C)}\end{tabular}} \\
\midrule

\rowcolor[gray]{0.92} \multicolumn{5}{c}{\textit{Confidence-based}} \\ 
\midrule
LogLikelihood & \textbf{0.9085} & \underline{0.8044} & 0.6459 & \underline{0.6951} \\
LogRank & \underline{0.9038} & \textbf{0.8138} & 0.6514 & \textbf{0.7010} \\
LRR & 0.2544 & 0.3698 & 0.4425 & 0.4081 \\
Entropy & 0.8655 & 0.4322 & 0.4701 & 0.4759 \\
\midrule

\rowcolor[gray]{0.92} \multicolumn{5}{c}{\textit{Fluctuation-based}} \\ 
\midrule
SpecDetect & 0.8411 & 0.7532 & \textbf{0.7017} & 0.5512 \\
\midrule

\rowcolor[gray]{0.92} \multicolumn{5}{c}{\textit{Fusion Indicator}} \\ 
\midrule
SpecFusion & 0.8998 & 0.7988 & \underline{0.6806} & 0.6497 \\
\bottomrule
\end{tabular}%
}
\end{table}

\paragraph{Confidence Superiority in Fragmented Contexts}
Sentence-level units give spectral methods short observation windows, so pure-generation rows favor confidence metrics: LogLikelihood outperforms SpecDetect on SemEval ($0.9085$ vs. $0.8411$), and LogRank leads on CoAuthor H vs. L ($0.8138$). This matches length sensitivity in Section~\ref{sec:exp_length}: confidence metrics exploit mean-probability shifts, while fluctuation indicators need longer spans to accumulate variance evidence.

\paragraph{Blending Effect in Collaborative Detection}
Collaborative rows are harder because they blend signatures from both sources: in CoAuthor, LogLikelihood drops from $0.8044$ on H vs. L to $0.6459$ on H vs. C, and LogRank falls from $0.8138$ to $0.6514$. Collaborative writing acts as a statistical buffer, weakening the clean low-perplexity footprint of machine text while disrupting fully human fluctuation patterns.

\paragraph{Stability vs. Optimality in Fusion}
SpecFusion provides a reliable safety net but does not always reach the best single-indicator ceiling. In H-vs-C, it maintains a stable AUC of $0.6806$, while SpecDetect reaches $0.7017$, suggesting that adaptive strategies should weigh confidence and fluctuation by text length and mixing type.




\subsection{RQ 2: Performance in Collaborative Text}
\label{sec:scenario_collab}


Beyond structural splicing, MixText lets us dissect how collaborative operations transform scores. We use pairwise accuracy, measuring whether a detector assigns a higher machine-origin score to the version with stronger machine involvement. Table~\ref{tab:mixtext_pairwise} reports results, Figure~\ref{fig:polish_viz} visualizes AI-Polishing signal changes, and the Humanizing visualization is in Appendix~\ref{app:mixtext_density}. The Llama3-8B proxy version is reported in Appendix~\ref{app:llama3_table4}.

\begin{table*}[t]
\centering
\caption{\textbf{Pairwise Accuracy on MixText Scenarios.} Values are the percentage of pairs where the Modified version is ranked more machine-like than the Original. Columns are ordered by machine involvement. Best results are \textbf{bolded}; second-best are \underline{underlined}.}
\label{tab:mixtext_pairwise}
\resizebox{0.85\textwidth}{!}{%
\begin{tabular}{l cccc cccc cccc}
\toprule
\multirow{3}{*}{\textbf{Metric}} & \multicolumn{4}{c}{\textbf{AI-Polishing (GPT-4)}} & \multicolumn{4}{c}{\textbf{AI-Polishing (Llama-2)}} & \multicolumn{4}{c}{\textbf{Humanizing}} \\
\cmidrule(lr){2-5} \cmidrule(lr){6-9} \cmidrule(lr){10-13}
 & \textbf{Polish} & \textbf{Polish} & \textbf{Complete} & \textbf{Rewrite} & \textbf{Polish} & \textbf{Polish} & \textbf{Complete} & \textbf{Rewrite} & \textbf{Adapt} & \textbf{Adapt} & \textbf{Llama-2} & \textbf{GPT-4} \\
 & \textbf{Token} & \textbf{Sentence} & \textbf{(1/3+2/3)} & \textbf{(Re-gen)} & \textbf{Token} & \textbf{Sentence} & \textbf{(1/3+2/3)} & \textbf{(Re-gen)} & \textbf{Token} & \textbf{Sentence} & \textbf{Humanize} & \textbf{Humanize} \\
\midrule
\rowcolor[gray]{0.92} \multicolumn{13}{c}{\textit{Confidence-based Metrics}} \\ 
\midrule
LogLikelihood & 0.6500 & 0.6167 & 0.6500 & 0.4067 & \underline{0.7433} & \underline{0.8233} & 0.8100 & \underline{0.9533} & \underline{0.7267} & \underline{0.8300} & \textbf{0.7333} & \textbf{0.9667} \\
LogRank & 0.6033 & 0.5800 & 0.6667 & 0.4233 & 0.6967 & \textbf{0.8300} & 0.7967 & \textbf{0.9600} & 0.7133 & \underline{0.8300} & \textbf{0.7333} & \underline{0.9600} \\
LRR & 0.3067 & 0.3133 & 0.5300 & \textbf{0.4933} & 0.3067 & 0.2700 & 0.1733 & 0.2767 & 0.2100 & 0.2167 & 0.2900 & 0.0800 \\
Entropy & \textbf{0.7867} & \textbf{0.7067} & 0.5067 & \underline{0.4867} & \textbf{0.8233} & 0.8167 & 0.8300 & 0.8300 & \textbf{0.7733} & 0.7600 & 0.6967 & \underline{0.9600} \\
\midrule
\rowcolor[gray]{0.92} \multicolumn{13}{c}{\textit{Fluctuation-based Metrics}} \\ 
\midrule
SpecDetect & 0.5267 & 0.5967 & \textbf{0.8100} & 0.4733 & 0.6400 & 0.7967 & \textbf{0.9700} & 0.7967 & 0.6767 & 0.7700 & \underline{0.7067} & 0.8833 \\
\midrule
\rowcolor[gray]{0.92} \multicolumn{13}{c}{\textit{Hybrid Fusion Indicators}} \\ 
\midrule
Lastde & \underline{0.6533} & \underline{0.6567} & 0.6967 & 0.4600 & 0.7367 & \textbf{0.8300} & \textbf{0.9700} & 0.8200 & 0.7200 & \textbf{0.8467} & \textbf{0.7333} & \textbf{0.9667} \\
SpecFusion & 0.6033 & 0.6233 & \underline{0.7100} & 0.4300 & 0.7333 & 0.8167 & \underline{0.9633} & 0.8367 & 0.6967 & 0.8033 & \textbf{0.7333} & 0.9333 \\
\bottomrule
\end{tabular}%
}
\end{table*}

A further edit-density diagnostic shows that surface edit amount alone has limited explanatory power for MixText behavior; operation type, edit direction, and continuity are more informative. Full definitions and task-level summaries are provided in Appendix~\ref{app:mixtext_density}.

\subsubsection{AI-Polishing: From Local Repairs to Global Recasting}
\label{sec:polish_analysis}

\begin{figure}[t]
    \centering
    \includegraphics[width=\linewidth]{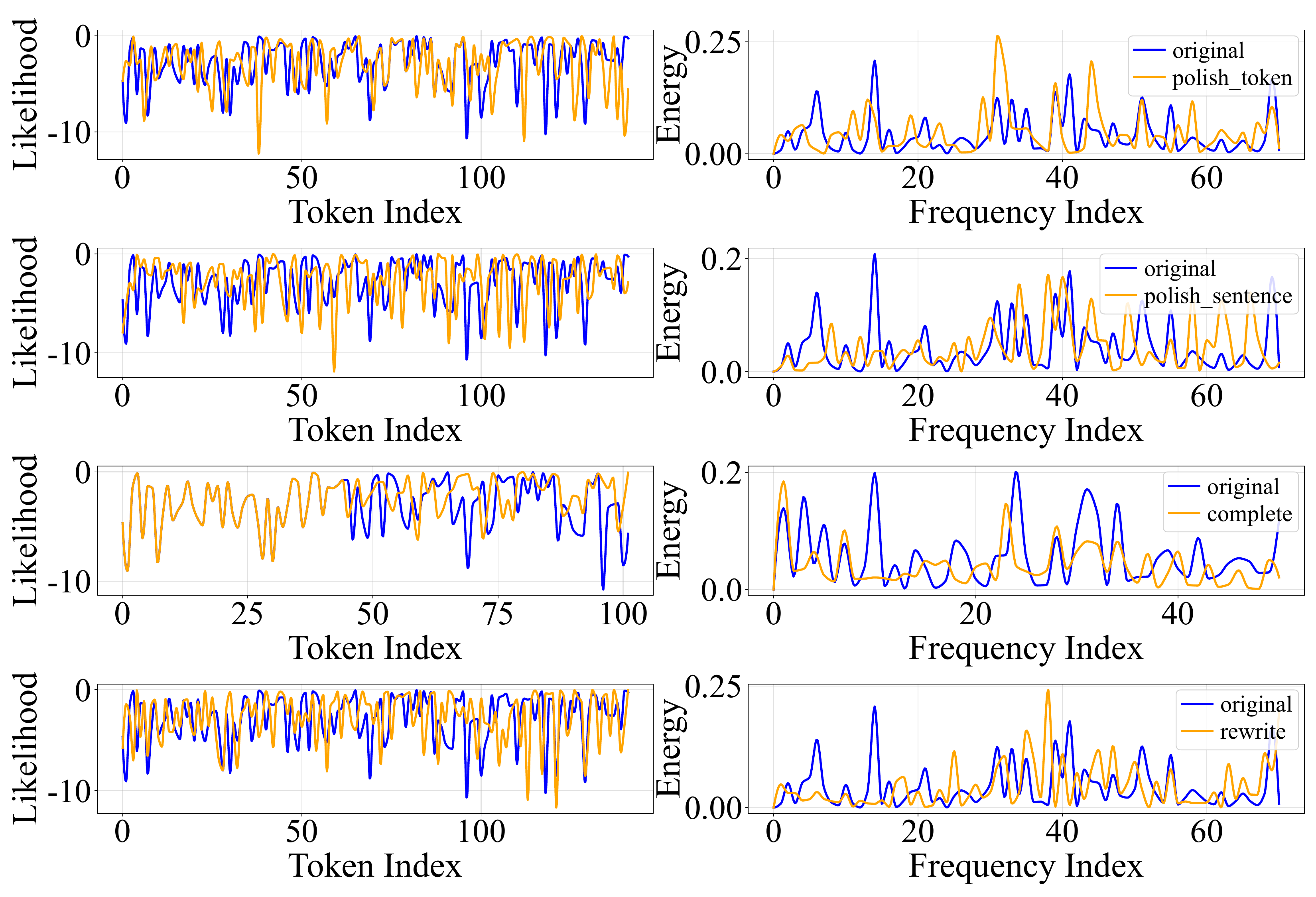} 
    \caption{\textbf{Case Visualization of Signal Profiles across AI-Polishing Modes.} Original and polished signals under each mode. Left: time-domain log-likelihood; right: frequency-domain spectrum.}
    \label{fig:polish_viz}
\end{figure}
Table~\ref{tab:mixtext_pairwise} and Figure~\ref{fig:polish_viz} reveal regimes governed by the continuity of LLM involvement.

\paragraph{Spectral Blindness in Point-wise Polishing}
Sparse token- or sentence-level polishing raises local confidence while leaving the underlying human fluctuation mostly intact. Confidence metrics capture this uplift: in GPT-4 Polish Token, Entropy reaches $78.67\%$ and LogLikelihood reaches $65.00\%$, while SpecDetect remains near chance at $52.67\%$. This source-inertia effect occurs because selected high-probability replacements rarely recast the sequence rhythm, so spectral signatures stay close to the human source.

\paragraph{Continuity and Stochasticity in Global Recasting}
When AI involvement expands into a continuous span, spectral evidence returns: in Complete, SpecDetect rises to $81.00\%$ on GPT-4 and $97.00\%$ on Llama-2 as autoregressive generation accumulates a low-variance machine pattern. Rewrite exposes the boundary, with Llama-2 Rewrite highly detectable by confidence metrics (LogRank $96.00\%$) while GPT-4 Rewrite is close to random for most metrics. Stable recasting is detectable, whereas high-entropy rewriting can reintroduce human-like roughness and weaken frequency-domain separation.

\subsubsection{Humanizing: Injection and Destructive Disguise}
\label{sec:humanize_analysis}

Table~\ref{tab:mixtext_pairwise} shows two strategies for altering machine signals in the AI $\to$ Human direction. The signal visualization is provided in Appendix~\ref{app:mixtext_density}.

\paragraph{Mimicking Roughness via Conservative Adaptation}
Manual adaptation differentiates machine text by adding human-like fluctuations: SpecDetect rises from $67.67\%$ on Adapt Token to $77.00\%$ on Adapt Sentence. Broader intervention widens the gap from the smooth machine baseline by reintroducing human-authorship ``grit'' into adapted text.

\paragraph{Detectability of Destructive Disguise}
Humanize operations inject artificial perturbations such as typos or unnatural errors. These manipulations degrade text naturalness while disrupting training-free metrics with minimal effort, exposing detectors' reliance on statistical signatures over semantic coherence. The paradox where text becomes ``less human'' yet ``less detectable'' reveals a fragility that warrants further attention.

\subsection{Concluding Takeaways} 
The in-the-wild evaluation yields four points. \textbf{Scale Sensitivity}: short or fragmented units favor confidence indicators because spectral evidence has little context. \textbf{Locality vs. Continuity}: point-wise polishing is captured by local probability shifts, whereas continuous completion gives spectral indicators a coherent generated span. \textbf{The Deception Paradox}: humanizing and rewriting can distort confidence and fluctuation in different ways, so edit density is insufficient. \textbf{Metric Complementarity}: confidence and fluctuation provide complementary safety nets across mixed and edited text.

\section{Conclusion}

This study establishes a physical foundation for LLM detection by reframing it as a frequency-domain signal processing task. We reveal that AI's suppression of linguistic variance creates distinct spectral signatures that dissipate under highly stochastic sampling. Spectral energy is strongest in long-form documents, while fragmented, mixed-source, and polished texts expose complementary confidence cues. Overall, our work explains the operating principles of spectral analysis and maps the conditions under which confidence and fluctuation evidence become useful for future multi-dimensional detection.
\section*{Limitations}
This work studies training-free detection through probability signals produced by a proxy language model. The resulting scores may vary with the choice of proxy model, decoding strategy, and text granularity, especially when the input contains very short spans or substantial human--AI editing. Our experiments cover standard document-level settings, mixed-source text, collaborative editing, additional proxy models, and corss-language tests, but broader deployment may involve languages, domains, generators, or editing styles beyond those evaluated here.

\bibliography{reference}

\clearpage
\section*{Appendix}
\appendix

\section{Extended Related Work}
\label{app:extended_related_work}

\subsection{Training-Free LLM Text Detection}
Training-free detection identifies machine-generated text via intrinsic statistical signatures, categorized into distribution-based approaches~\cite{mitchell2023detectgpt,su2023detectllm,yang2023dna,bao2023fast} using perturbations and sample-based methods~\cite{solaiman2019release,su2023detectllm,gehrmann2019gltr,ippolito2019automatic,xu2024training,luo2025specdetect} utilizing raw token statistics. We focus on the latter as they represent fundamental machine signal properties. Early research relied on confidence-based metrics like LogLikelihood, LogRank~\cite{solaiman2019release}, and Entropy~\cite{gehrmann2019gltr,ippolito2019automatic}, assuming LLMs exhibit higher statistical certainty than humans. Subsequent work like Lastde~\cite{xu2024training} began modeling probability volatility, leading to the significant advancement of SpecDetect~\cite{luo2025specdetect}. SpecDetect reframes detection as a signal processing task by analyzing fluctuations via the Discrete Fourier Transform (DFT). Despite its effectiveness, the theoretical mechanisms and physical properties of frequency-domain detection remain opaque, a gap this study aims to bridge.

\subsection{Detection in Complex Scenarios}
Digital content increasingly involves complex scenarios like mixed-source text~\cite{lee2022coauthor,wang2024semeval} and collaborative authorship~\cite{zhang2024llm}, where interleaved segments and human-AI editing alter the text's statistical traces. While recent studies~\cite{zeng2024detecting,su2025haco,kadiyala2025robust} employ supervised models for sequence labeling or boundary detection, they often suffer from limited cross-domain generalization~\cite{zeng2024detecting}. Conversely, training-free detection remains confined to long-form, single-source documents, leaving its performance in intricate scenarios unexplored. Although frequency-domain methods such as SpecDetect~\cite{luo2025specdetect} effectively capture generative fluctuations at the document level, it remains unknown whether indicators can maintain discriminative power when signals are fragmented by mixing or masked by collaborative editing. This work interrogates these complex regimes to define the utility and limitations of frequency-domain detection beyond the traditional long-form, single-source paradigm.

\section{Mathematical Proofs}

\subsection{Derivation of Theorem \ref{thm:variance}}
\label{app:proof_variance}

\begin{proof}
For a text source $s$ and token position $t$, the indicator $b_t^{(s)}$ records whether the token falls in $\mathcal{V}_{\mathrm{tail}}^{(t)}(\tau)$, and $\gamma_s$ is the average rate at which this event occurs. Applying the law of total variance~\cite{yu2021assessment} to the mixture signal $x_s[t]$ gives
\begin{equation}
    \operatorname{Var}(x_s[t])
    =
    \mathbb{E}[\operatorname{Var}(x_s[t]\mid b_t^{(s)})]
    +
    \operatorname{Var}(\mathbb{E}[x_s[t]\mid b_t^{(s)}]).
\end{equation}
The first term combines the conditional variances inside the head and tail regions:
\begin{equation}
    \mathbb{E}[\operatorname{Var}(x_s[t]\mid b_t^{(s)})]
    =
    (1-\gamma_s)\sigma_{\mathrm{head}}^2
    +\gamma_s\sigma_{\mathrm{tail}}^2 .
\end{equation}
The second term captures the variance caused by switching between the head and tail means:
\begin{equation}
    \operatorname{Var}(\mathbb{E}[x_s[t]\mid b_t^{(s)}])
    =
    \gamma_s(1-\gamma_s)(\mu_{\mathrm{tail}}-\mu_{\mathrm{head}})^2 .
\end{equation}
Combining the two terms gives the group-wise variance
\begin{align}
    \sigma_s^2
    &=
    (1-\gamma_s)\sigma_{\mathrm{head}}^2
    +\gamma_s\sigma_{\mathrm{tail}}^2 \nonumber\\
    &\quad+
    \gamma_s(1-\gamma_s)(\mu_{\mathrm{tail}}-\mu_{\mathrm{head}})^2 .
\end{align}
Substituting $s=H$ and $s=\mathrm{AI}$ and subtracting $\sigma_{\mathrm{AI}}^2$ from $\sigma_H^2$ yields
\begin{align}
    \Delta\sigma^2
    &=
    (\gamma_H-\gamma_{\mathrm{AI}})
    (\sigma_{\mathrm{tail}}^2-\sigma_{\mathrm{head}}^2) \nonumber\\
    &\quad+
    \bigl[\gamma_H(1-\gamma_H) \nonumber\\
    &\qquad
    -\gamma_{\mathrm{AI}}(1-\gamma_{\mathrm{AI}})\bigr]
    (\mu_{\mathrm{tail}}-\mu_{\mathrm{head}})^2 .
\end{align}
Additionally, in the idealized white-box truncation case where generated tokens remain in the head region, setting $\gamma_{\mathrm{AI}}=0$ gives
\begin{align}
    \Delta\sigma^2
    =
    \gamma_H(\sigma_{\mathrm{tail}}^2-\sigma_{\mathrm{head}}^2) \nonumber\\
    +\gamma_H(1-\gamma_H)
    (\mu_{\mathrm{tail}}-\mu_{\mathrm{head}})^2 .
\end{align}
The human--machine fluctuation gap therefore emerges when the tail rate and conditional log-probability structure differ across text sources. Tail-region log-probabilities are typically lower and more dispersed, so a higher human tail-token rate increases the variance gap under the stated shared-statistics view.
\end{proof}

\subsection{Proof of Corollary~\ref{thm:energy}}
\label{app:proof_energy}

\begin{proof}
For a text source $s\in\{H,\mathrm{AI}\}$, let $z_s[t]=x_s[t]-\mu_s$ be the centered scoring-model log-probability sequence of length $N$. We use the Discrete Fourier Transform (DFT)~\cite{bracewell1989fourier},
\begin{equation}
    \hat{z}^{(s)}_k = \sum_{t=0}^{N-1} z_s[t] e^{-j\frac{2\pi}{N}kt}.
\end{equation}

Parseval's theorem~\cite{hardy1931note} for this convention states:
\begin{equation}
    \sum_{k=0}^{N-1}|\hat{z}^{(s)}_k|^2 = N\sum_{t=0}^{N-1}z_s[t]^2 .
\end{equation}

We define the average spectral energy as
\begin{equation}
    \bar{\mathcal{E}}_s
    =
    \frac{1}{N}\sum_{k=0}^{N-1}|\hat{z}^{(s)}_k|^2 .
\end{equation}
Substituting Parseval's identity into this definition gives
\begin{equation}
    \bar{\mathcal{E}}_s
    =
    \sum_{t=0}^{N-1}z_s[t]^2 .
\end{equation}
The variance of the centered signal is
\begin{equation}
    \sigma_s^2
    =
    \frac{1}{N}\sum_{t=0}^{N-1}z_s[t]^2,
    \quad
    \sum_{t=0}^{N-1}z_s[t]^2
    =
    N\sigma_s^2 .
\end{equation}
Therefore,
\begin{equation}
    \bar{\mathcal{E}}_s
    =
    N\sigma_s^2 .
\end{equation}

Taking expectations for human and AI text sources and subtracting gives
\begin{align}
    \Delta\bar{\mathcal{E}}
    &=
    \mathbb{E}[\bar{\mathcal{E}}_H]
    -
    \mathbb{E}[\bar{\mathcal{E}}_{\mathrm{AI}}] \nonumber\\
    &=
    N(\sigma_H^2-\sigma_{\mathrm{AI}}^2)
    =
    N\Delta\sigma^2 .
\end{align}
Thus, the spectral-energy gap follows directly from the time-domain fluctuation divergence established in Theorem~\ref{thm:variance}.
\end{proof}

\section{Experimental Setup Details}
\subsection{Metric Details}
\label{app:metric_details}

In this section, we provide the formal definitions for all detection metrics. For consistency, all metrics are formulated as a score function $\mathcal{S}$ where a \textbf{higher value indicates a higher probability of machine generation.}

\subsubsection{Confidence-based Metrics}
These metrics measure the average probability intensity and capture the mean shift inherent in AI's restricted sampling.

\begin{itemize}[leftmargin=*]
    \item \textbf{LogLikelihood (LL)}~\cite{solaiman2019release}: 
    The primary measure of model certainty. Since AI models maximize likelihood, machine text yields higher values.
    \begin{equation}
        \mathcal{S}_{\text{LL}} = \frac{1}{N} \sum_{t=1}^{N} \log P(w_t | w_{<t})
    \end{equation}

    \item \textbf{LogRank (LR)}~\cite{solaiman2019release}: 
    The rank-domain counterpart to Log-Likelihood. It measures the usage of high-probability (low-rank) tokens. To align polarity, we use the negative logarithmic rank:
    \begin{equation}
        \mathcal{S}_{\text{LR}} = - \frac{1}{N} \sum_{t=1}^{N} \log(\text{Rank}(w_t))
    \end{equation}

    \item \textbf{LRR (Log-Likelihood Ratio)}~\cite{su2023detectllm}: 
    A normalized metric that balances probability magnitude against rank magnitude. It is defined directly as the ratio of Log-Likelihood to the average Log-Rank:
    \begin{equation}
        \mathcal{S}_{\text{LRR}} = \frac{\text{LogLikelihood}}{\text{LogRank}} = \frac{\frac{1}{N} \sum_{t=1}^{N} \log P(w_t)}{\frac{1}{N} \sum_{t=1}^{N} \log(\text{Rank}(w_t))}
    \end{equation}

    \item \textbf{Entropy}~\cite{gehrmann2019gltr,ippolito2019automatic}: 
    Captures predictive uncertainty. AI text exhibits low entropy (high certainty). We define the score as the negative average entropy:
    \begin{equation}
        \mathcal{S}_{\text{Entropy}} = - \frac{1}{N} \sum_{t=1}^{N} \mathbb{H}(P(\cdot|w_{t}))
    \end{equation}
\end{itemize}

\subsubsection{Fluctuation-based Metrics}
These metrics measure the structural variance or spectral density of the signal.

\begin{itemize}[leftmargin=*]
    

    \item \textbf{SpecDetect}~\cite{luo2025specdetect}: 
    Integrates the spectral energy of the log-probability signal. Physically, AI text lacks high amplitude fluctuations, leading to low spectral energy. To eliminate the impact of inconsistent sequence lengths across candidate samples, we adopt the average spectral energy.
    
    Let the original time-domain signal be $x[t] = \log P(w_t | w_{<t})$. We first perform zero-centering to remove the mean component:
    \begin{equation}
        \hat{x}[t] = x[t] - \frac{1}{N}\sum_{j=1}^{N} x[j].
    \end{equation}
    We then define the score as the negative average power of the DFT coefficients $X[k]$:
    \begin{equation}
        \mathcal{S}_{\text{SpecDetect}} = - \frac{1}{N} \sum_{k=0}^{N-1} |X[k]|^2,
    \end{equation}
    where  $X[k] = \sum_{t=0}^{N-1} \hat{x}[t] \cdot e^{-i \frac{2\pi}{N} kt}$.
\end{itemize}

\subsubsection{Fusion-based Metrics}
These indicators combine the mean (confidence) and variance (fluctuation) dimensions to bridge the blind spots of single-view metrics.

\begin{itemize}[leftmargin=*]
    \item \textbf{Lastde}~\cite{xu2024training}: 
    This method models fluctuation using Multiscale Diversity Entropy (MDE). The score is defined as the ratio of likelihood to fluctuation. It reaches its maximum when a signal is simultaneously high-confidence and low-fluctuation:
    \begin{equation}
        \mathcal{S}_{\text{LastDe}} = \frac{\mathcal{S}_{\text{LL}}}{\text{MDE}}
    \end{equation}
    where MDE denotes the magnitude of Multiscale Diversity Entropy computed on the rank sequence. Adopting the default configuration, we set the hyperparameters as follows:  $s=3$,  $\epsilon=10\times N$, and $\tau^{'}=5$, with $N$ representing the number of tokens in the text.

    \item \textbf{SpecFusion}: 
    A standardized linear combination. Since both $\mathcal{S}_{\text{LL}}$ and $\mathcal{S}_{\text{SpecDetect}}$ are aligned such that higher values indicate AI, we sum their standardized scores:
    \begin{equation}
        \mathcal{S}_{\text{Fusion}} = Z(\mathcal{S}_{\text{LL}}) + Z(\mathcal{S}_{\text{SpecDetect}})
    \end{equation}
    where $Z(\cdot)$ denotes Z-score standardization. The normalization scope $\mathcal{C}$ depends on the task granularity:
    \begin{itemize}
        \item For document-level detection, $\mathcal{C}$ includes all documents in the evaluation corpus.
        \item For sentence-level detection, $\mathcal{C}$ includes all sentences within the current document, effectively normalizing against the local context.
    \end{itemize}
\end{itemize}

\subsection{Dataset Details}
\label{app:dataset_details}

To evaluate detection performance across different granularities of human-AI interaction, the experiments are categorized into two primary real-world scenarios. Table~\ref{tab:dataset_stats} summarizes the statistics of the datasets utilized in this study. The selected benchmarks cover diverse detection levels (i.e., sentence-level and document-level) and originate from various LLM generators (e.g., GPT-3/4~\cite{achiam2023gpt}, Llama-2~\cite{touvron2023llama2}), ensuring a comprehensive assessment of metric robustness under varying signal densities.

\begin{table}[t]
\centering
\caption{\textbf{Dataset Statistics.} $Avg.\ Len.$ is the word count per detection unit. $Mixed$ denotes mixed-source text within a document; $Collab$ indicates collaborative content.}
\label{tab:dataset_stats}
\resizebox{\columnwidth}{!}{%
\begin{tabular}{l l l c c c c c}
\toprule
\textbf{Dataset} & \textbf{Sub-task} & \textbf{LLM Engine} & \textbf{Samples} & \textbf{Avg. Len.} & \textbf{Level} & \textbf{Mixed} & \textbf{Collab} \\
\midrule
SemEval & Prefix-Comp. & GPT-3 & 4,154 & 19.76 & Sent. & \checkmark & $\times$ \\
CoAuthor & Interleaved & GPT-3.5 & 1,445 & 14.71 & Sent. & \checkmark & \checkmark \\
\midrule
\multirow{2}{*}{MixText} & AI-Polishing & GPT-4, Llama-2 & 3,000 & 130.18 & Doc. & $\times$ & \checkmark \\
 & Humanizing & GPT-4, Llama-2 & 1,500 & 128.66 & Doc. & $\times$ & \checkmark \\
\bottomrule
\end{tabular}%
}
\end{table}

\paragraph{Scenario I: Mixed-Source Text.} 
This scenario targets documents where human-authored and AI-generated sentences coexist, reflecting common ``spliced'' or ``interleaved'' content structures found in news co-writing or academic stitching.
\begin{itemize}[leftmargin=*]
    \item \textbf{SemEval-2024 (Task 8)}~\cite{zhang2024llm}: The experiments utilize Subtask C from this benchmark. The data consists of documents with a human-written prefix followed by a machine-generated continuation (GPT-3). Since the transition point is variable, this task evaluates the capability to identify machine segments within a composite context.
    \item \textbf{CoAuthor}~\cite{lee2022coauthor}: Moving beyond the simple prefix-continuation structure, CoAuthor simulates a complex environment where sentences are interleaved throughout the text. Each document contains sentences from three sources: \textbf{Human}, \textbf{LLM}, and \textbf{Collaborative}. This setup serves to evaluate detection performance in non-sequential mixing environments.
\end{itemize}

\paragraph{Scenario II: Collaborative Text.} 
Complementing the structural mixing in CoAuthor, the \textbf{MixText}~\cite{zhang2024llm} dataset is employed to conduct a granular investigation into specific editing operations. The dataset comprises 300 sample pairs for each modification direction, generated using GPT-4 and Llama-2-70b engines.
\begin{itemize}[leftmargin=*]
    \item \textbf{AI-Polishing (Human $\to$ AI)}: Machine-led refinement of human drafts. This category includes three specific modes:
    \begin{itemize}
        \item \textit{Polish}: Involves \textbf{Token-level} and \textbf{Sentence-level} refinement. In this mode, the LLM focuses on enhancing fluency and correcting minor errors while strictly adhering to the original semantic skeleton.
        \item \textit{Rewrite}: The LLM extracts key information and regenerates the text entirely. Unlike simple polishing, this operation significantly alters the syntactic structure.
        \item \textit{Complete}: The LLM generates the remaining 2/3 of the text based on a human-authored 1/3 prefix, representing a continuous autoregressive generation process.
    \end{itemize}
    
    \item \textbf{Humanizing (AI $\to$ Human)}: Strategies used to disguise machine origin, categorized into two opposing approaches:
    \begin{itemize}
        \item \textit{Humanize}: A ``destructive'' strategy that introduces artificial noise (e.g., typos, grammatical errors) to mimic human writing imperfections. This injection process is specifically implemented using prompt-engineering on GPT-4 and Llama-2 models.
        \item \textit{Adapt}: A ``conservative'' strategy where humans align machine text with natural linguistic habits to enhance fluency. Similar to the polishing task, this operation is further stratified into \textbf{Token-level} and \textbf{Sentence-level} adaptations to verify detection sensitivity to edit granularity.
    \end{itemize}
\end{itemize}

\subsection{Implementation Details}
\label{app:real_protocol}

Following the configurations established in our document-level experiments, we utilize \textbf{GPT-J-6B} as the base proxy model to extract all statistical metrics. To address the unique challenges posed by diverse real-world human-AI interactions, we adopt two distinct evaluation protocols:
\begin{itemize}[leftmargin=*]
    \item \textbf{Scenario I: Mixed-Source Text Evaluation}. 
    This protocol measures the ability to distinguish between authorship sources within a single, heterogeneous document context. We perform sentence-level scoring across the document and report the average AUC calculated within each individual sample. This approach ensures that the detection performance reflects the metric's sensitivity to fragmented signals and local statistical shifts within each document.
    
    \item \textbf{Scenario II: Collaborative Text Evaluation}. 
    Focusing on the MixText dataset, we employ pairwise accuracy to assess detection performance. This metric calculates the percentage of samples where the metric score correctly shifts following a specific editing operation. For instance, a successful detection is recorded when an AI-polished version of a text receives a higher detection score than its original human-authored counterpart. This methodology directly measures the sensitivity of each indicator to the fine-grained statistical transformations introduced by collaborative editing and humanization processes.
\end{itemize}

All experiments are conducted on a high-performance computing node equipped with a single 80GB NVIDIA H800 GPU. To ensure the reproducibility of our results, our implementation builds upon the open-source frameworks provided by the original authors. The complete codebase and configurations are available at the following repositories:
\begin{itemize}[leftmargin=*]
    \item \textbf{Other metrics}: \url{https://github.com/TrustMedia-zju/Lastde_Detector}
    \item \textbf{SpecDetect}: \url{https://github.com/luohaitong/SpecDetect}
\end{itemize}

\section{Additional Experiment Results}
\label{app:additional_experiments}

\subsection{Metric Distinctness Validation}
\label{app:metric_validation}

To validate the generality of our findings, we extend the dimensionality analysis to the XSum and Reddit datasets.

\begin{figure}
    \centering
    \subfigure[Results on XSum Dataset]{
        \includegraphics[width=\linewidth]{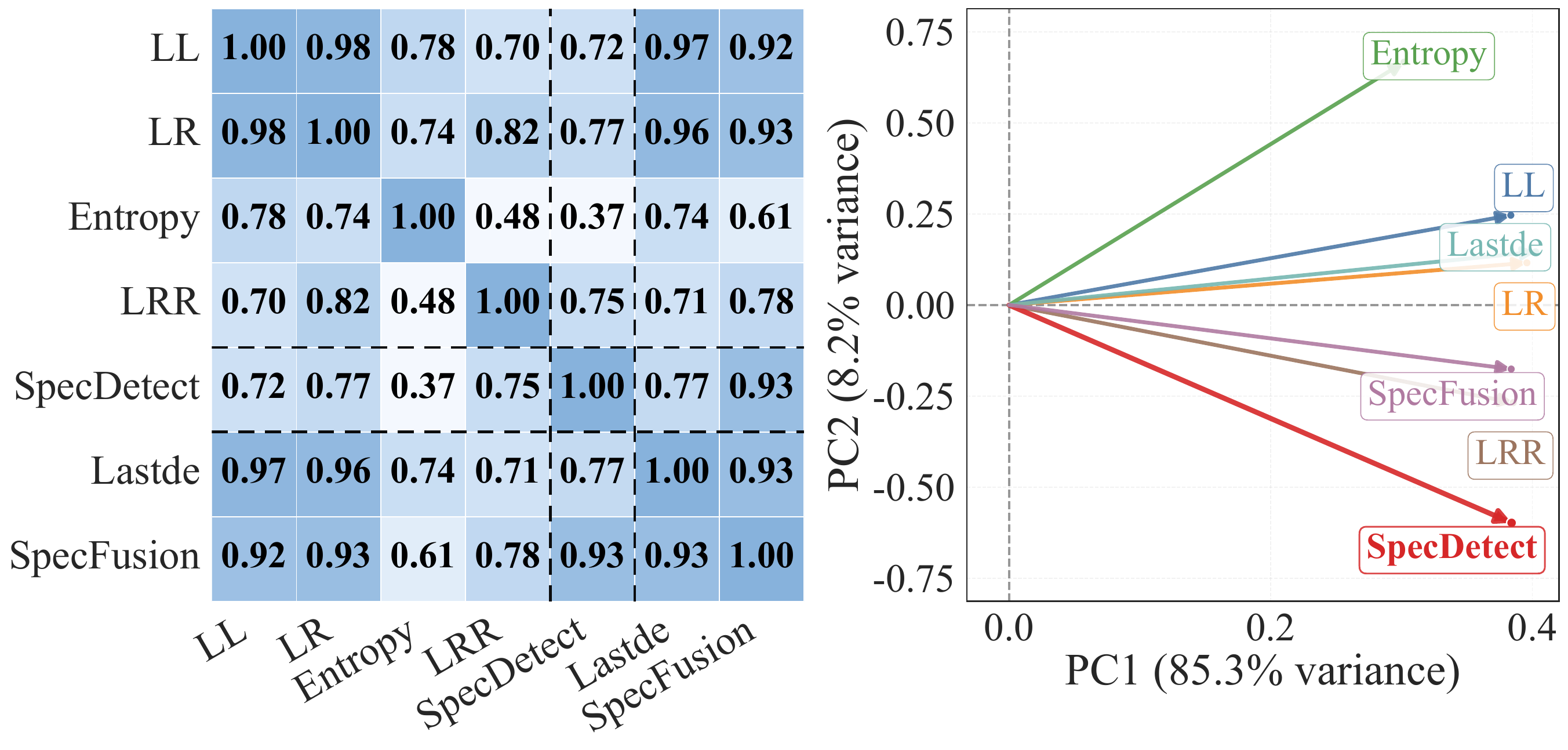}
    }
    \subfigure[Results on Reddit Dataset]{
        \includegraphics[width=\linewidth]{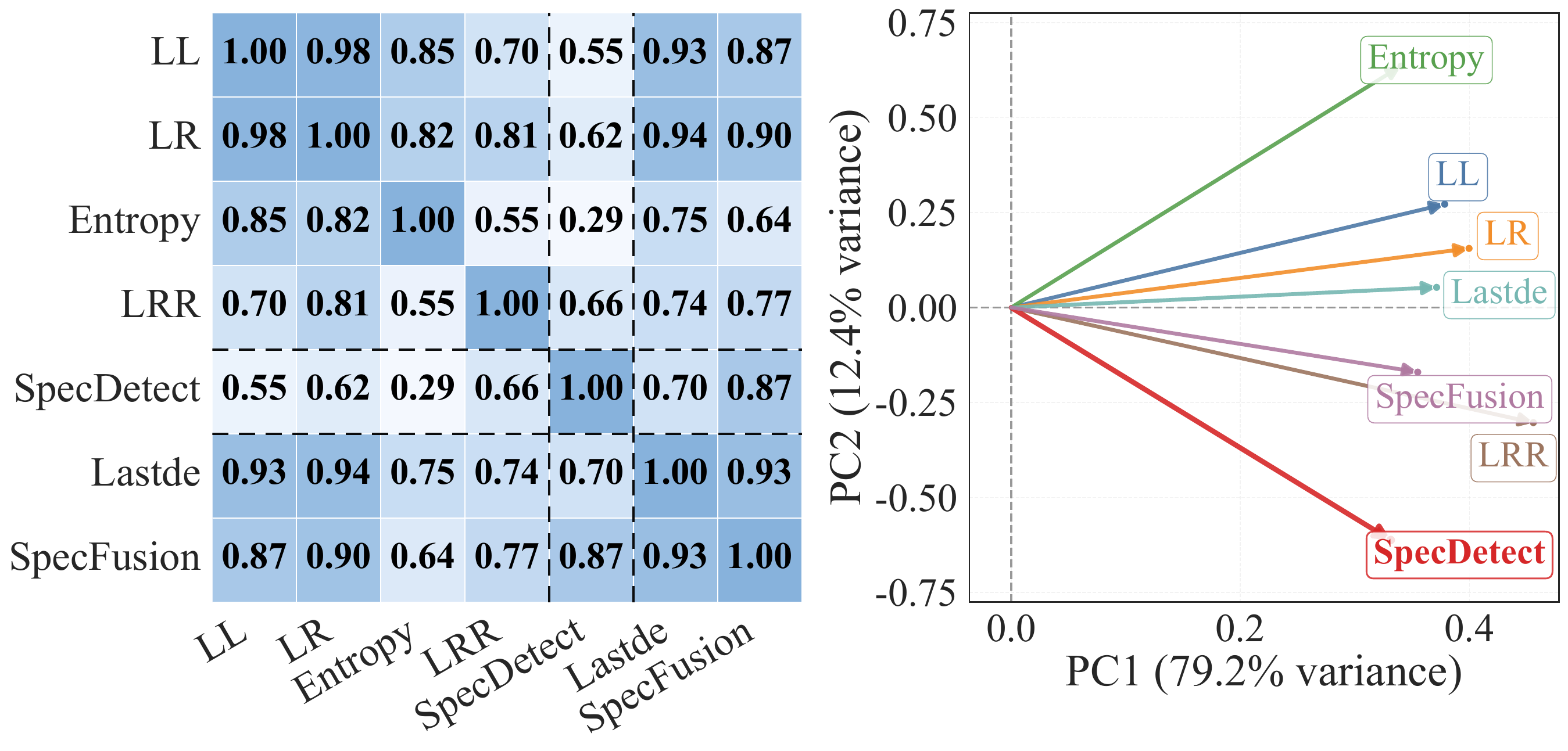}
    }
    \caption{\textbf{Analysis of Metric Relation on Additional Datasets.} 
    (Left) \textbf{Correlation Heatmap}: Metrics cluster into Confidence and Fluctuation families based on Spearman correlation. 
    (Right) \textbf{PCA Loading Plot}: Illustrates the geometric relationship where PC1 captures the dominant variance, while PC2 captures the secondary direction of variation.}
    \label{fig:app_metric_analysis}
\end{figure}

As shown in Figure~\ref{fig:app_metric_analysis}, the results are consistent with the Writing dataset results presented in the main text. The correlation heatmaps confirm that metrics cluster tightly within their respective families. Similarly, the PCA loading plots demonstrate that spectral methods capture a distinct signal dimension separate from the confidence-based probability.

\subsection{Input Signal Granularity Investigation}
\label{app:ablation_proxy}

To capture spectral signatures accurately, the choice of the input proxy signal $x[t]$ is critical. While prior work~\cite{mitchell2023detectgpt,bao2023fast} demonstrates that LogRank provides a robust, scale-invariant measure for Confidence-based metrics, whether this preference holds for fluctuation-based indicators remains unexamined.

\begin{figure}
    \centering
    \includegraphics[width=\linewidth]{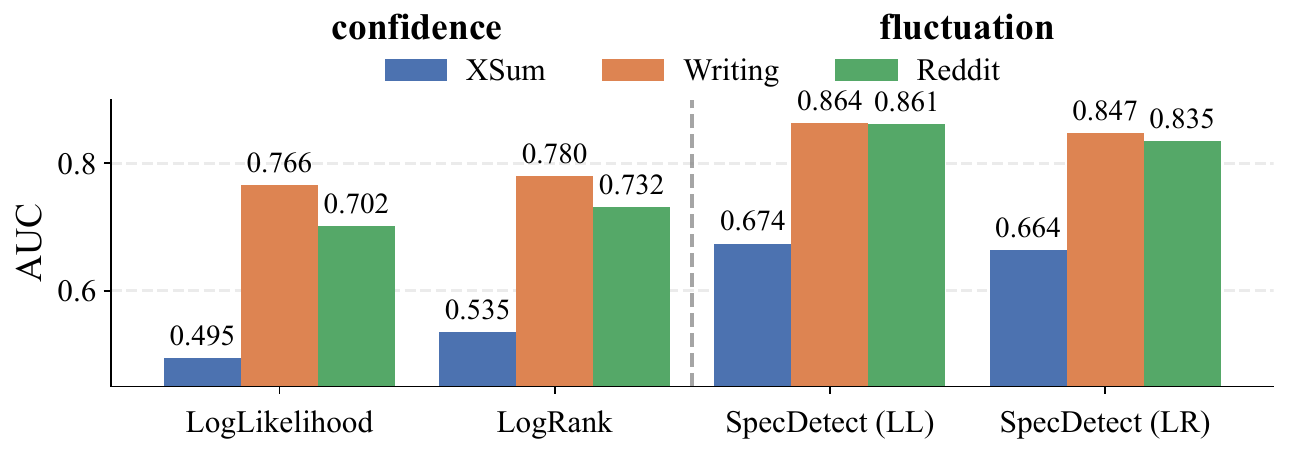}
    \caption{\textbf{Input Proxy Ablation across All Metrics.} While confidence metrics benefit from discretized LogRank, fluctuation-based metrics require continuous \textsf{LogLikelihood} to resolve high-frequency spectral signatures.
    Comparative analysis of input signal preference. While confidence metrics improve with discretized LogRank inputs, fluctuation-based metrics (i.e., SpecDetect) suffer a severe performance collapse, confirming the necessity of continuous LogLikelihood for resolving micro-fluctuations.}
    \label{fig:proxy_ablation}
\end{figure}

As illustrated in Figure~\ref{fig:proxy_ablation}, we observe a distinct divergence in signal preference across three datasets. While confidence-based metrics improve by $+2\%\sim4\%$ AUC with LogRank, fluctuation-based metrics suffer a performance collapse, achieving optimal results \textit{only} with continuous LogLikelihood.

This phenomenon stems from Information Granularity. Physically, LogRank acts as a non-linear quantizer that discretizes the probability manifold. While this quantization filters noise for coarse mean estimation (benefiting confidence metrics), it effectively smooths out the fine-grained micro-fluctuations and rhythmic pulses. Since spectral analysis relies on these variations to resolve the variance gap, the quantization noise destroys the signal's structural integrity. Thus, LogLikelihood is essential for fluctuation-based detection, whereas LogRank remains optimal for confidence-based baselines.

\subsection{Complete Main-Setting Sensitivity Results}
\label{app:main_setting_sensitivity}

This subsection complements the main setting, where Llama2-13B is the source model and GPT-J-6B is the proxy scorer. The main text highlights representative metrics and datasets; here we report fuller curves for length and sampling sensitivity.

\subsubsection{Length Sensitivity}
\label{app:sensitivity_length}

\begin{figure}
    \centering
    \includegraphics[width=1\linewidth]{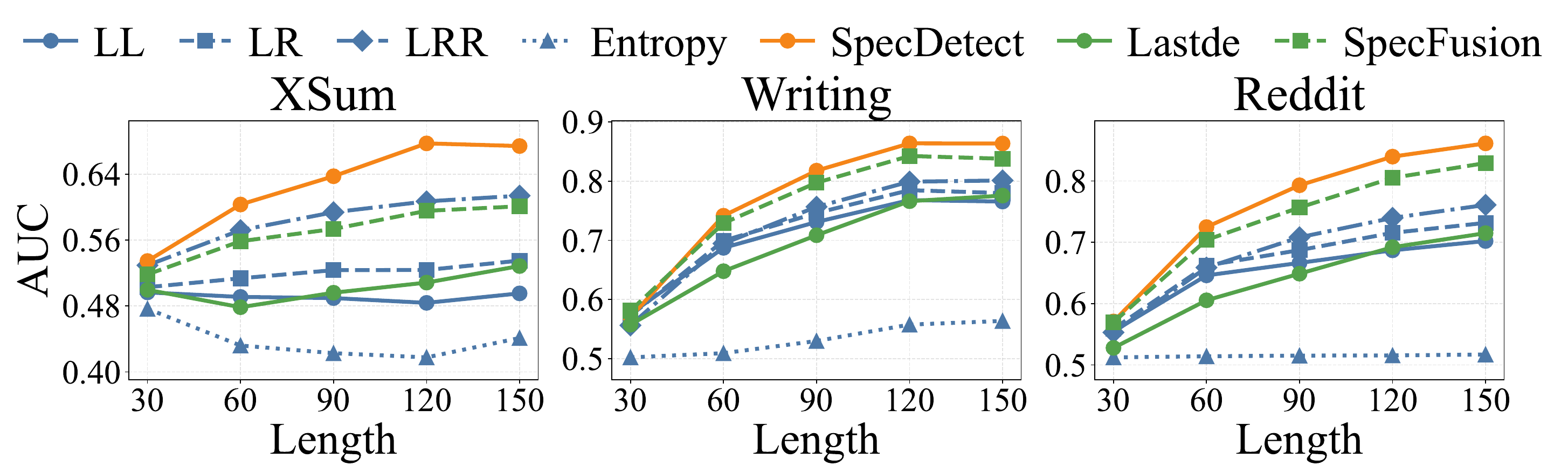} 
    \caption{\textbf{Performance of All Metrics vs. Sequence Length (Llama-2 Source).} Fluctuation-based metrics achieve more significant gains as sequence length increases.}
    \label{fig:app_length_llama2}
\end{figure}

Figure~\ref{fig:app_length_llama2} reports the all-metric length curves under the main source/proxy setting. The result is consistent with Figure~\ref{fig:length_impact}: fluctuation-based metrics gain more clearly as sequence length increases, because longer spans provide more evidence for stable variance patterns.

\subsubsection{Sampling Scope}
\label{app:sensitivity_sampling}

\begin{figure}
    \centering
    \subfigure[Results on XSum Dataset]{
        \includegraphics[width=\linewidth]{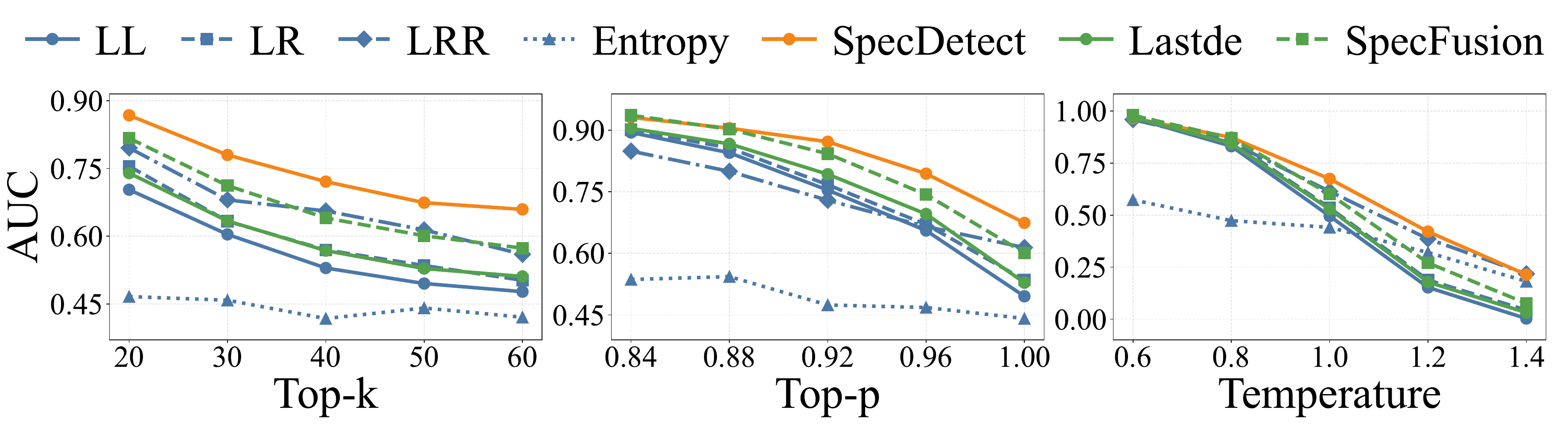}
    }
    \subfigure[Results on writing Dataset]{
        \includegraphics[width=\linewidth]{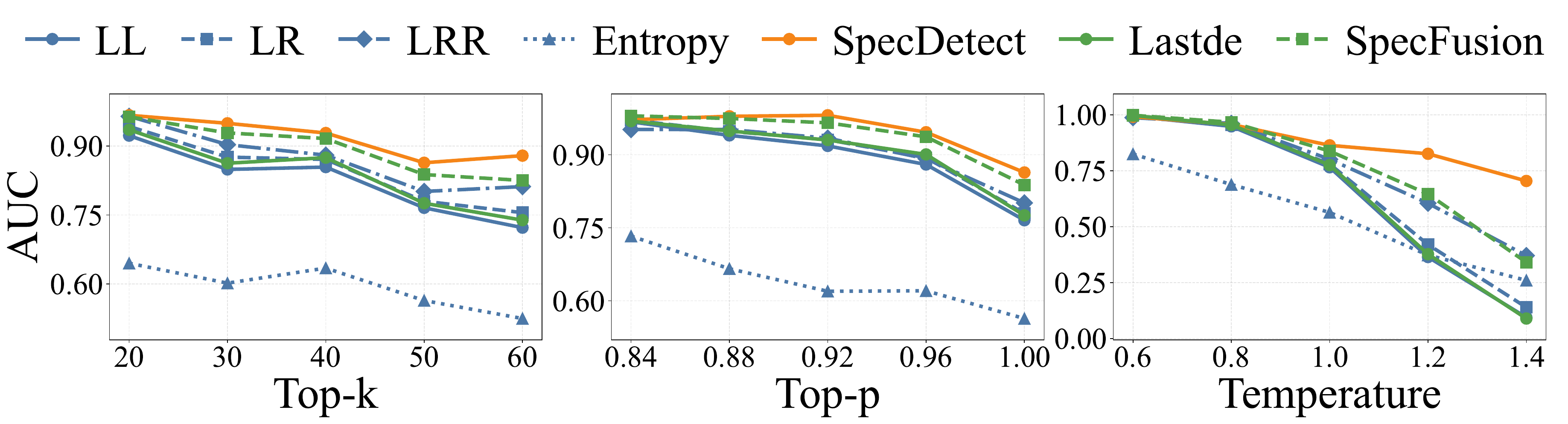}
    }
    \subfigure[Results on Reddit Dataset]{
        \includegraphics[width=\linewidth]{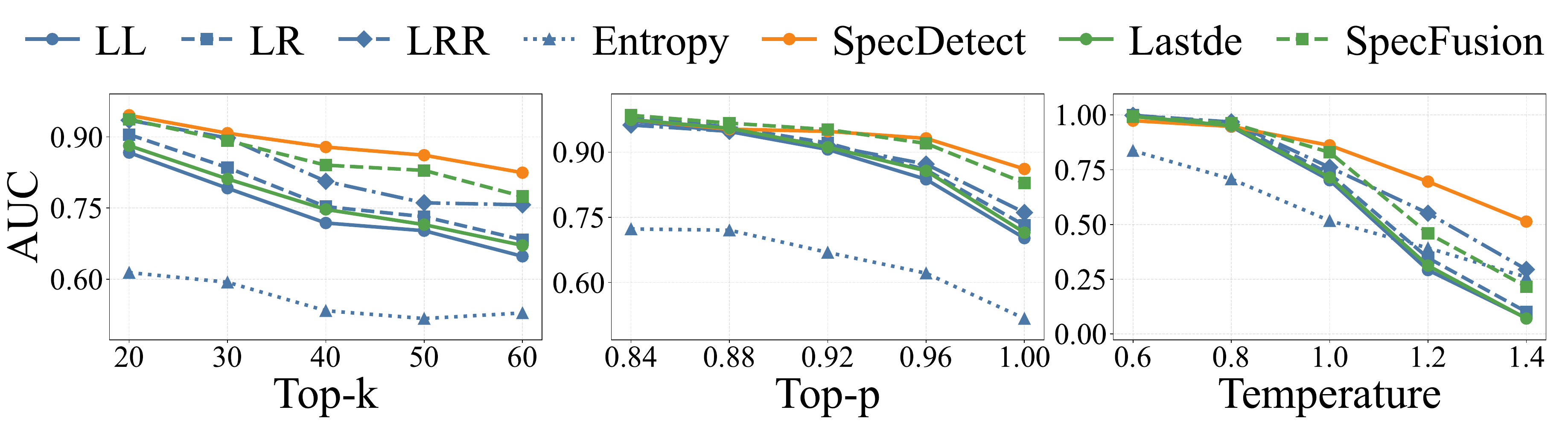}
    }
    \caption{\textbf{Impact of Decoding Strategies across All Datasets (Llama-2 Source).} Results across three evaluation scenarios demonstrate that tighter sampling constraints consistently enhance detection performance for all metrics, reinforcing the physical consistency of the variance collapse phenomenon.}
    \label{fig:app_decoding_llama2}
\end{figure}

Figure~\ref{fig:app_decoding_llama2} reports the all-dataset sampling-scope curves under the main source/proxy setting. The pattern follows Figure~\ref{fig:full_sensitivity}: broader sampling generally weakens separability, while tighter decoding makes the machine signal easier to distinguish.

\subsection{Additional Source-Model Sensitivity Results}
\label{app:source_model_sensitivity}

This subsection checks whether the length and sampling-scope observations persist when the source model changes. We use GPT-4-Turbo for the additional length study and Qwen-3-8B for the additional sampling-scope study. The results are used as qualitative source-model checks, since the exact AUC values and slopes naturally depend on the generator.

\subsubsection{Length Sensitivity}
\label{app:source_length}

\begin{figure}
    \centering
    \includegraphics[width=1\linewidth]{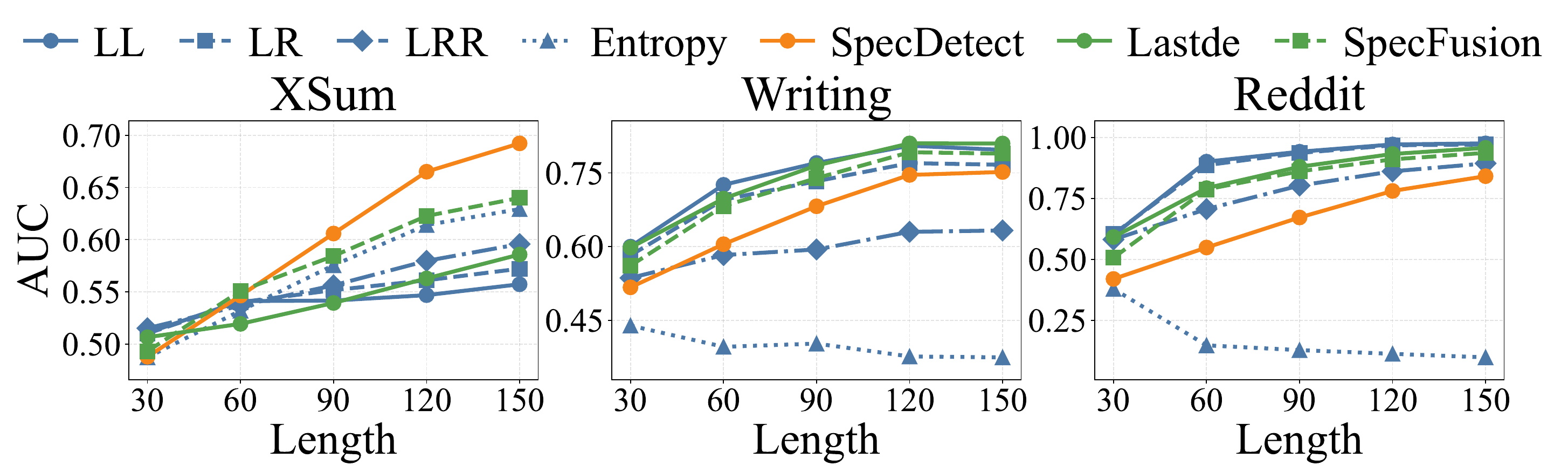} 
    \caption{\textbf{Performance of All Metrics vs. Sequence Length (GPT-4 Source).} While SpecDetect initially trails on Writing and Reddit, it achieves the steepest growth among all metrics as length increases, confirming that spectral methods require sufficient observation windows to resolve stable variance.}
    \label{fig:app_length_gpt4}
\end{figure}

Figure~\ref{fig:app_length_gpt4} shows the length-sensitivity curves for GPT-4-Turbo generated text. The trend remains aligned with the main setting: SpecDetect initially trails on some datasets but grows strongly with longer observation windows.

\subsubsection{Sampling Scope}
\label{app:source_sampling}

\begin{figure}
    \centering
    \subfigure[Results on XSum Dataset]{
        \includegraphics[width=\linewidth]{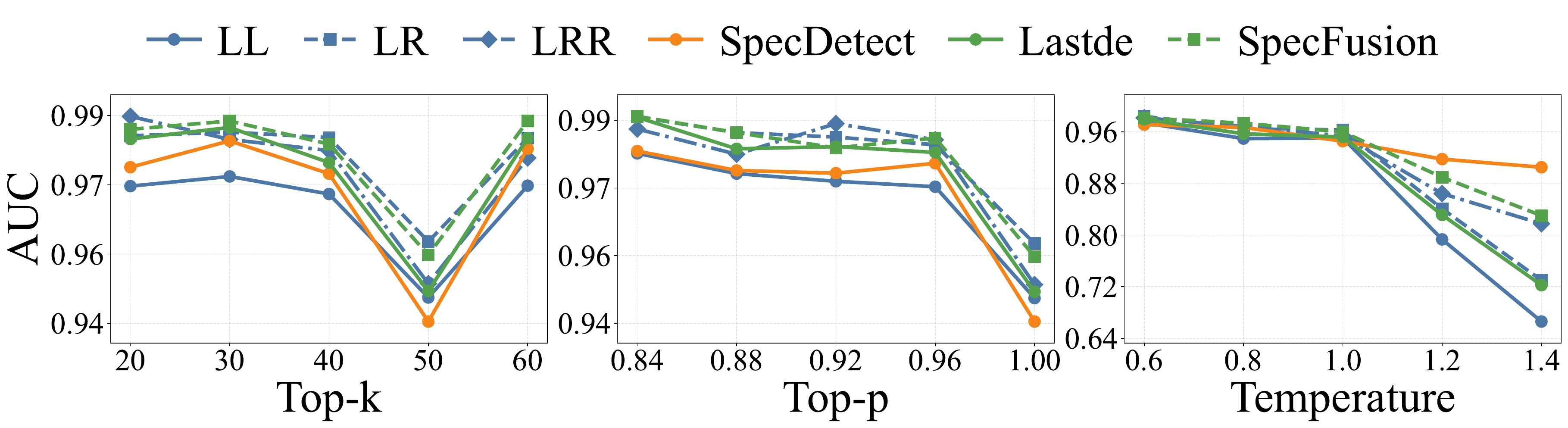}
    }
    \subfigure[Results on writing Dataset]{
        \includegraphics[width=\linewidth]{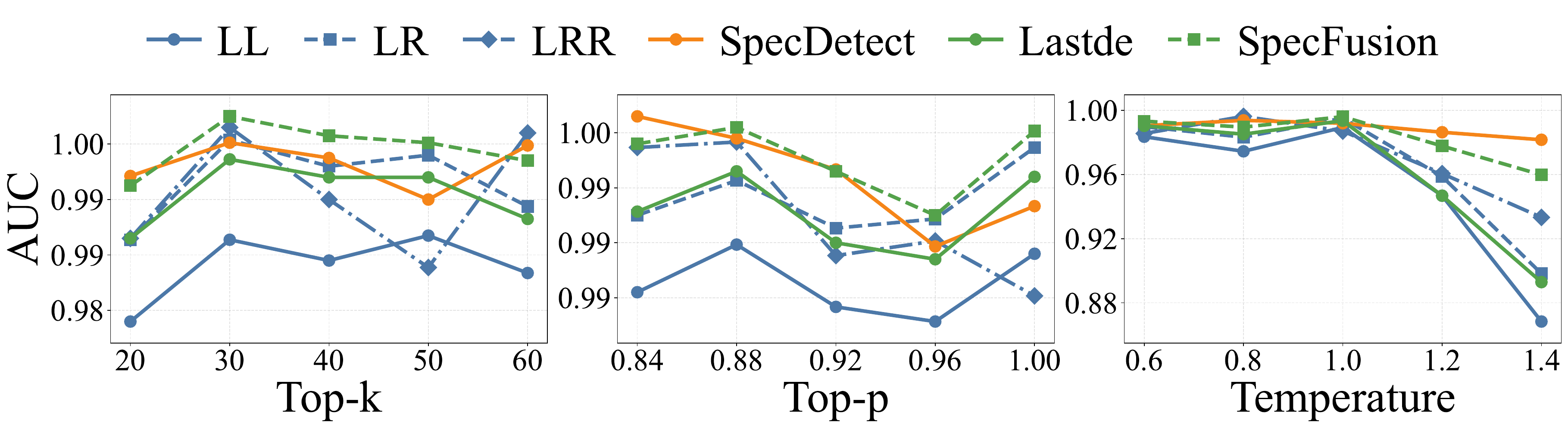}
    }
    \subfigure[Results on Reddit Dataset]{
        \includegraphics[width=\linewidth]{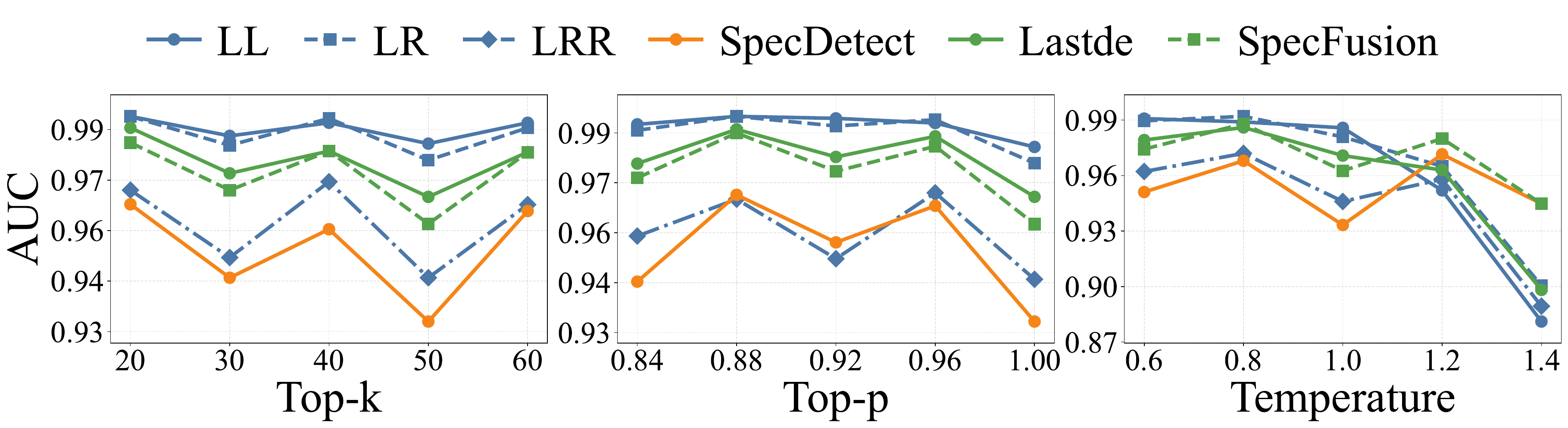}
    }
    \caption{\textbf{Impact of Decoding Strategies across All Datasets (Qwen-3-8B Source).} Tighter sampling improves separability across the three evaluation scenarios, while the exact gradients vary with the source model.}
    \label{fig:app_decoding_qwen3}
\end{figure}

Figure~\ref{fig:app_decoding_qwen3} shows the sampling-scope curves for Qwen-3-8B generated text. The performance gradient is more subtle under this source model, but the overall trajectories remain consistent with the main conclusion that stricter sampling improves separability. We exclude Entropy from these visualizations for clarity because its near-random performance otherwise obscures the more discriminative indicators.

\subsection{Additional Head/Tail Validation}
\label{app:headtail}

\begin{figure}[t]
    \centering
    \includegraphics[width=\linewidth]{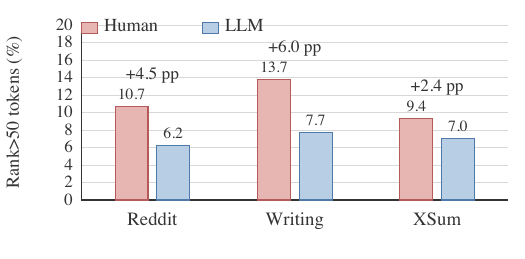}
    \caption{\textbf{GPT-J-6B Rank$>$50 head/tail check.} Bars show the percentage of tokens whose proxy-model rank is greater than 50.}
    \label{fig:headtail_gptj_rank50}
\end{figure}

\begin{figure}[t]
    \centering
    \includegraphics[width=\linewidth]{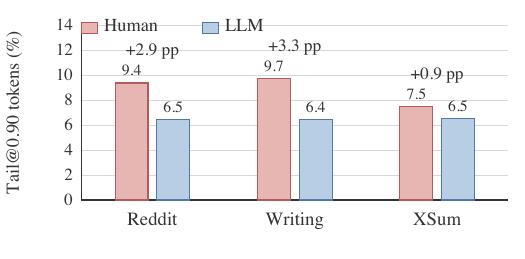}
    \caption{\textbf{Llama3-8B Tail@0.90 head/tail check.} Bars show the percentage of tokens outside the Top-$p=0.90$ head set under Llama3-8B proxy scoring.}
    \label{fig:headtail_llama3_tail90}
\end{figure}

\begin{figure}[t]
    \centering
    \includegraphics[width=\linewidth]{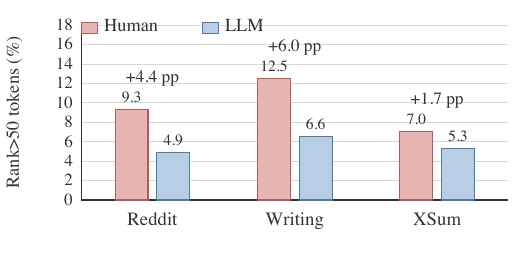}
    \caption{\textbf{Llama3-8B Rank$>$50 head/tail check.} Bars show the percentage of tokens whose proxy-model rank is greater than 50.}
    \label{fig:headtail_llama3_rank50}
\end{figure}

These additional head/tail checks are consistent with the main-text GPT-J-6B Tail@0.90 result in Figure~\ref{fig:headtail_prelim}. Figure~\ref{fig:headtail_gptj_rank50} uses GPT-J-6B with a Rank$>$50 threshold, while Figures~\ref{fig:headtail_llama3_tail90} and~\ref{fig:headtail_llama3_rank50} repeat the Tail@0.90 and Rank$>$50 checks using Llama3-8B. In all three views, human continuations enter the proxy tail region more often than paired AI continuations, with larger gaps on Reddit and WritingPrompts and a smaller gap on XSum.

\subsection{Llama3-8B Proxy Results for Main Experiments}
\label{app:llama3_proxy}

This section reports the available Llama3-8B scoring-proxy counterparts for the main experiments. These figures and tables complement the GPT-J-6B proxy results in the main text. Overall, the qualitative conclusions are consistent with the GPT-J-6B setting: longer spans strengthen fluctuation-based evidence, broader sampling weakens all detectors while preserving a relative fluctuation advantage in the evaluated range, and mixed or collaborative sentence-level settings remain harder than pure H-vs-L detection.

\subsubsection{Length Sensitivity}
\label{app:llama3_length}

\begin{figure}[t]
    \centering
    \includegraphics[width=\linewidth]{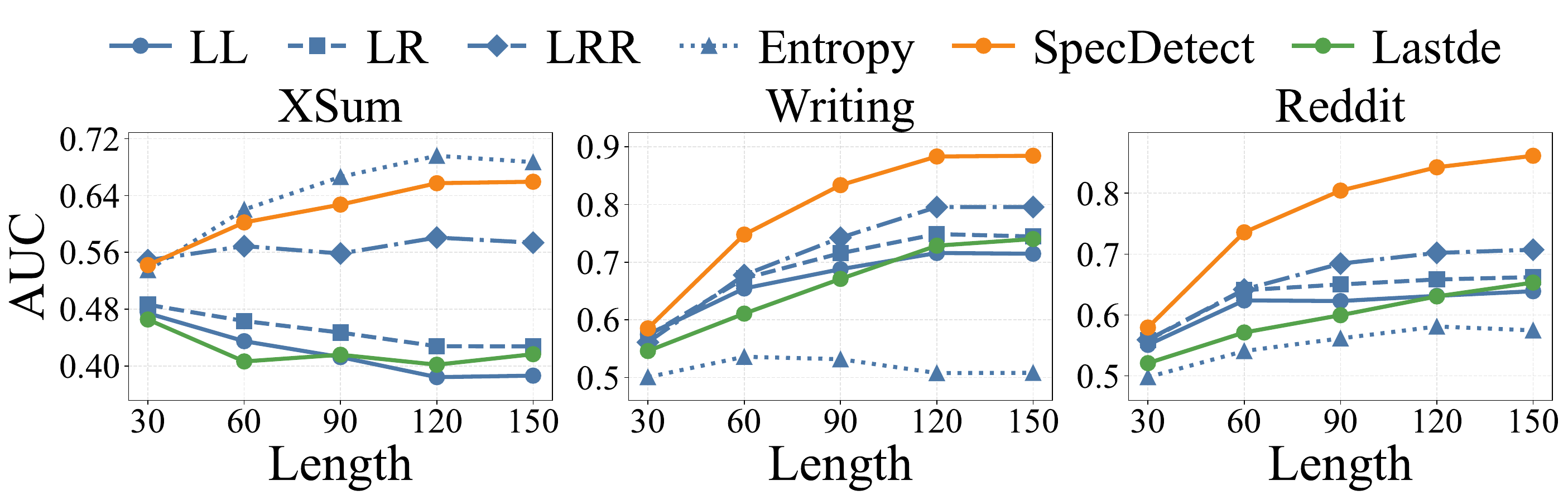}
    \caption{\textbf{Llama3-8B proxy length sensitivity.} This figure mirrors Figure~\ref{fig:length_impact} with Llama3-8B as the scoring proxy. The qualitative pattern is consistent with the main text: longer spans tend to strengthen fluctuation-based evidence, while exact values vary with the proxy model.}
    \label{fig:llama3_length_proxy}
\end{figure}

Figure~\ref{fig:llama3_length_proxy} shows the Llama3-8B proxy counterpart of the main length-sensitivity analysis. The qualitative trend is consistent with Figure~\ref{fig:length_impact}: longer spans usually strengthen fluctuation-based evidence, although the exact AUC values and slopes vary with the proxy model.

\subsubsection{Sampling Scope}
\label{app:llama3_sampling}

\begin{figure}[t]
    \centering
    \subfigure[Results on XSum Dataset]{
        \includegraphics[width=\linewidth]{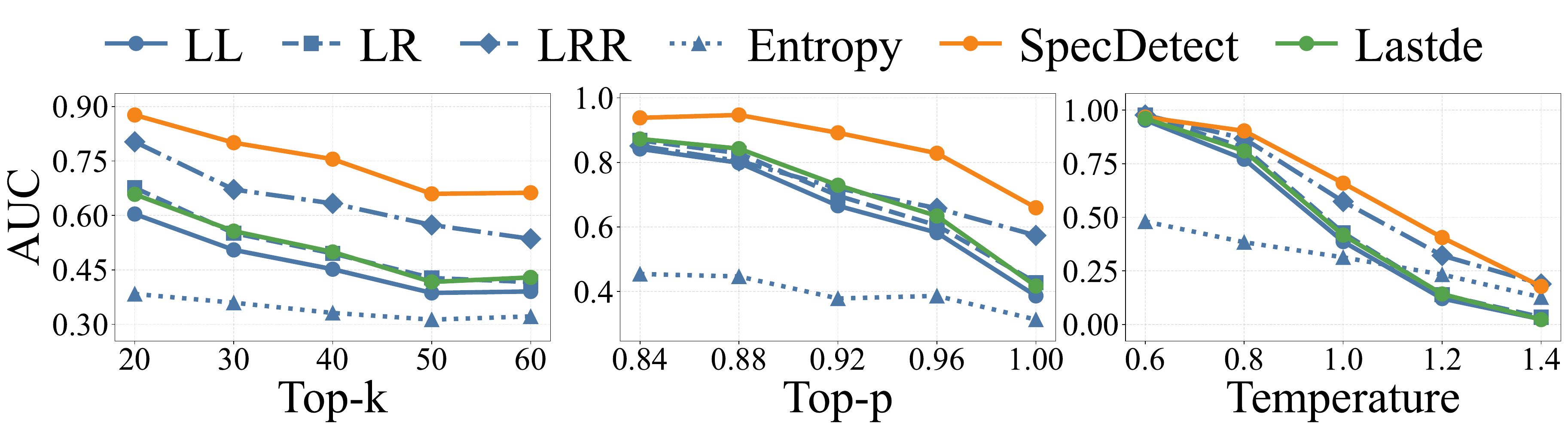}
    }
    \subfigure[Results on WritingPrompts Dataset]{
        \includegraphics[width=\linewidth]{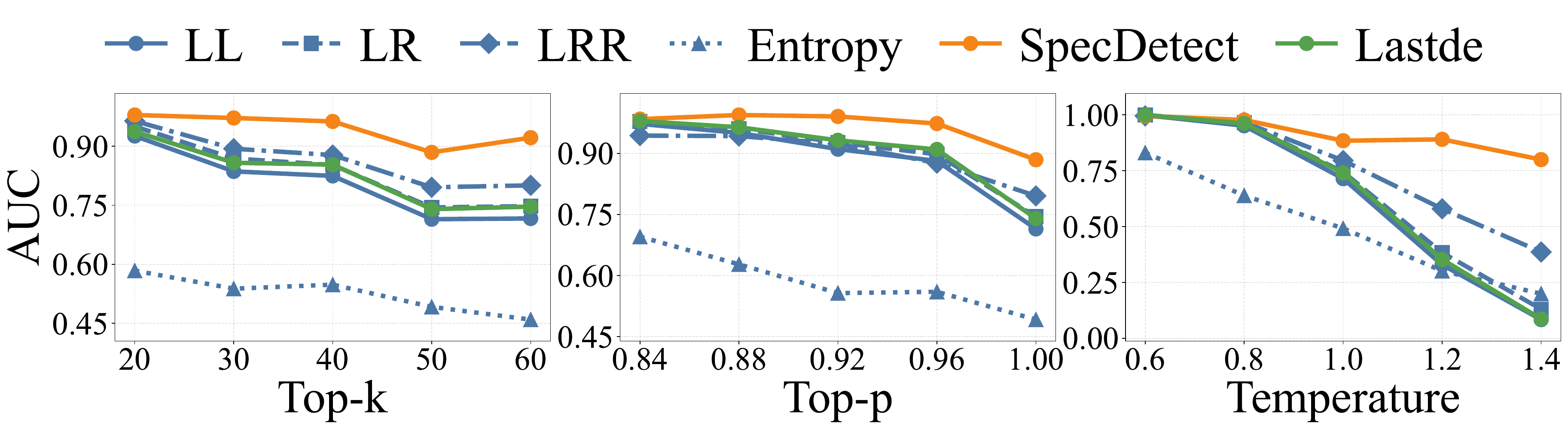}
    }
    \subfigure[Results on Reddit Dataset]{
        \includegraphics[width=\linewidth]{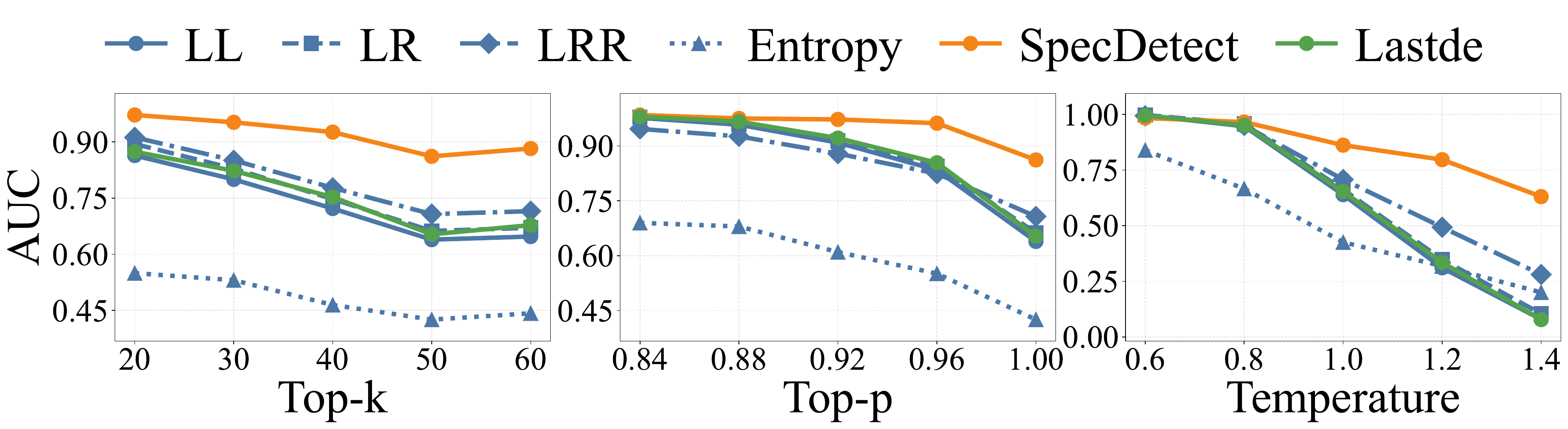}
    }
    \caption{\textbf{Impact of Decoding Strategies across All Datasets (Llama3-8B Proxy).} This figure mirrors the sampling-scope analysis with Llama3-8B as the scoring proxy. Broader sampling generally weakens separability, while exact curves vary with the proxy model and dataset.}
    \label{fig:llama3_sampling_datasets}
\end{figure}

Figure~\ref{fig:llama3_sampling_datasets} shows the Llama3-8B proxy counterpart of the sampling-scope analysis across XSum, WritingPrompts, and Reddit. The pattern is aligned with Figure~\ref{fig:full_sensitivity}: broader sampling generally lowers detector performance, while fluctuation-based indicators show relatively smoother degradation in the evaluated range.

\subsubsection{Sentence-Level Pure and Collaborative Text}
\label{app:llama3_table3}

Table~\ref{tab:llama3_table3} reports the Llama3-8B proxy counterpart of Table~\ref{tab:mixed_results}. The results follow the same broad pattern: confidence metrics are strong in pure H-vs-L sentence-level settings, while collaborative rows are harder and fluctuation evidence becomes more useful in the L-vs-C comparison.

\begin{table}[t]
\centering
\caption{\textbf{Llama3-8B Proxy Sentence-level Detection across Pure and Collaborative Text.} Average within-document AUC for binary classification among Human (H), LLM (L), and Collaborative (C) sentences, following the same format as Table~\ref{tab:mixed_results}. Best results are \textbf{bolded}; second-best are \underline{underlined}.}
\label{tab:llama3_table3}
\resizebox{0.85\columnwidth}{!}{%
\begin{tabular}{lcccc}
\toprule
\multicolumn{1}{c}{\multirow{2}{*}{\textbf{Metric}}} & \multicolumn{2}{c}{\textbf{Pure Generation}} & \multicolumn{2}{c}{\textbf{Collaborative Mixed}} \\
\cmidrule(lr){2-3} \cmidrule(lr){4-5}
 & \textbf{SemEval} & \multicolumn{1}{c}{\begin{tabular}{@{}c@{}}\textbf{CoAuthor}\\\textbf{(H vs. L)}\end{tabular}} & \multicolumn{1}{c}{\begin{tabular}{@{}c@{}}\textbf{CoAuthor}\\\textbf{(H vs. C)}\end{tabular}} & \multicolumn{1}{c}{\begin{tabular}{@{}c@{}}\textbf{CoAuthor}\\\textbf{(L vs. C)}\end{tabular}} \\
\midrule

\rowcolor[gray]{0.92} \multicolumn{5}{c}{\textit{Confidence-based}} \\ 
\midrule
LogLikelihood & \textbf{0.8698} & \underline{0.6415} & \underline{0.7083} & 0.5839 \\
LogRank & 0.8449 & \textbf{0.6483} & \textbf{0.7221} & 0.5982 \\
LRR & 0.2190 & 0.5152 & 0.5152 & 0.5276 \\
Entropy & 0.2981 & 0.5306 & 0.5306 & 0.4751 \\
\midrule

\rowcolor[gray]{0.92} \multicolumn{5}{c}{\textit{Fluctuation-based}} \\ 
\midrule
SpecDetect & 0.7141 & 0.5639 & 0.6721 & \textbf{0.7479} \\
\midrule

\rowcolor[gray]{0.92} \multicolumn{5}{c}{\textit{Fusion Indicator}} \\ 
\midrule
SpecFusion & \underline{0.8479} & 0.6412 & 0.6931 & \underline{0.7453} \\
\bottomrule
\end{tabular}}
\end{table}

\subsubsection{MixText Pairwise Accuracy}
\label{app:llama3_table4}

Table~\ref{tab:llama3_table4} reports the Llama3-8B proxy counterpart of Table~\ref{tab:mixtext_pairwise}. The operation-level pattern is similar to the main text: continuous completion favors fluctuation or hybrid indicators, while local polishing and humanizing operations often rely more on confidence shifts.

\begin{table*}[t]
\centering
\caption{\textbf{Llama3-8B Proxy Pairwise Accuracy on MixText Scenarios.} Values are the percentage of pairs where the Modified version is ranked more machine-like than the Original, following the same column order and formatting as Table~\ref{tab:mixtext_pairwise}. Best results are \textbf{bolded}; second-best are \underline{underlined}.}
\label{tab:llama3_table4}
\small
\resizebox{0.85\textwidth}{!}{%
\begin{tabular}{l cccc cccc cccc}
\toprule
\multirow{3}{*}{\textbf{Metric}} & \multicolumn{4}{c}{\textbf{AI-Polishing (GPT-4)}} & \multicolumn{4}{c}{\textbf{AI-Polishing (Llama-2)}} & \multicolumn{4}{c}{\textbf{Humanizing}} \\
\cmidrule(lr){2-5} \cmidrule(lr){6-9} \cmidrule(lr){10-13}
 & \textbf{Polish} & \textbf{Polish} & \textbf{Complete} & \textbf{Rewrite} & \textbf{Polish} & \textbf{Polish} & \textbf{Complete} & \textbf{Rewrite} & \textbf{Adapt} & \textbf{Adapt} & \textbf{Llama-2} & \textbf{GPT-4} \\
 & \textbf{Token} & \textbf{Sentence} & \textbf{(1/3+2/3)} & \textbf{(Re-gen)} & \textbf{Token} & \textbf{Sentence} & \textbf{(1/3+2/3)} & \textbf{(Re-gen)} & \textbf{Token} & \textbf{Sentence} & \textbf{Humanize} & \textbf{Humanize} \\
\midrule
\rowcolor[gray]{0.92} \multicolumn{13}{c}{\textit{Confidence-based Metrics}} \\ 
\midrule
LogLikelihood & 0.3733 & \textbf{0.6267} & 0.5667 & 0.5300 & 0.6933 & 0.8133 & 0.9633 & \underline{0.7733} & \textbf{0.7600} & \textbf{0.8833} & \textbf{0.7267} & \textbf{0.9900} \\
LogRank & 0.5033 & 0.5167 & 0.6467 & 0.3767 & 0.6833 & 0.7867 & \underline{0.9767} & 0.7567 & 0.7200 & \underline{0.8600} & \underline{0.7100} & 0.9733 \\
LRR & 0.3200 & 0.3367 & 0.5100 & \underline{0.5333} & 0.3133 & 0.2100 & 0.1067 & 0.2600 & 0.1467 & 0.1267 & 0.2967 & 0.0433 \\
Entropy & 0.3533 & 0.4300 & 0.5567 & \textbf{0.6233} & 0.2567 & 0.2100 & 0.1133 & 0.2100 & 0.1667 & 0.1600 & 0.2433 & 0.0300 \\
\midrule
\rowcolor[gray]{0.92} \multicolumn{13}{c}{\textit{Fluctuation-based Metrics}} \\ 
\midrule
SpecDetect & \textbf{0.5633} & \underline{0.6133} & \textbf{0.8033} & 0.4900 & 0.6933 & 0.8100 & 0.9600 & 0.7633 & 0.6967 & 0.8433 & 0.6600 & 0.9067 \\
\midrule
\rowcolor[gray]{0.92} \multicolumn{13}{c}{\textit{Hybrid Fusion Indicators}} \\ 
\midrule
Lastde & 0.5500 & 0.5933 & 0.6633 & 0.3933 & \textbf{0.7233} & \underline{0.8200} & \textbf{0.9800} & \textbf{0.8033} & \underline{0.7433} & 0.8500 & 0.6933 & \underline{0.9833} \\
SpecFusion & \underline{0.5533} & 0.5833 & \underline{0.7433} & 0.3967 & \underline{0.7100} & \textbf{0.8267} & \textbf{0.9800} & \underline{0.7733} & 0.7200 & \underline{0.8600} & 0.6967 & 0.9600 \\
\bottomrule
\end{tabular}}
\end{table*}

\subsection{Cross-Language Test}
\label{app:crosslingual_domain}

This appendix reports a WMT16~\cite{bojar2016findings}(German) cross-language test for the two boundary conclusions in the main text. We provide a length-sensitivity view and a decoding-scope view following the main protocol.

\begin{figure}[t]
    \centering
    \includegraphics[width=0.7\linewidth]{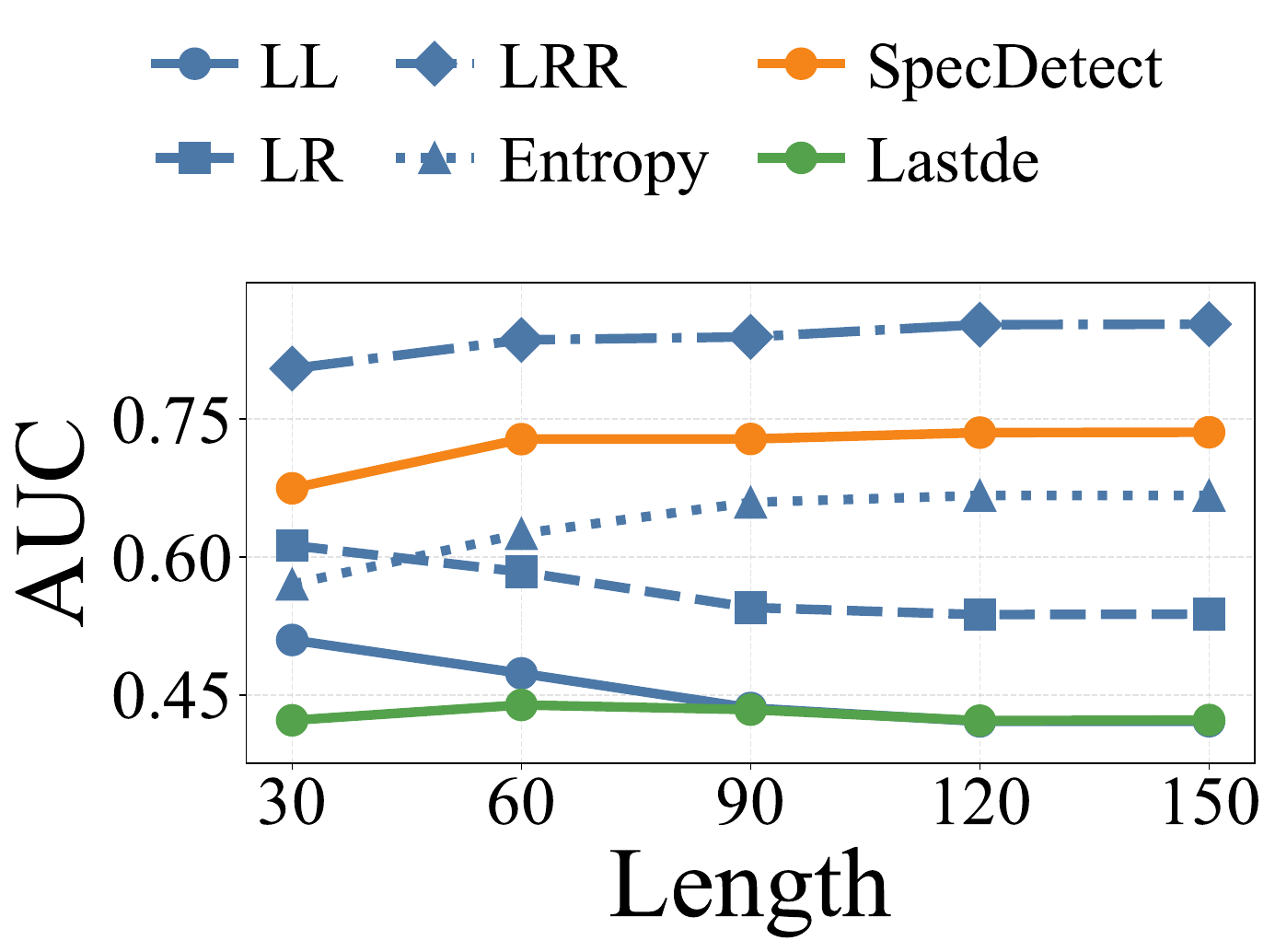}
    \caption{\textbf{WMT16 cross-language test (length sensitivity).} Following Figure~\ref{fig:length_impact}.}
    \label{fig:crosslingual_wmt16_length}
\end{figure}
\begin{figure}[t]
    \centering
    \includegraphics[width=\linewidth]{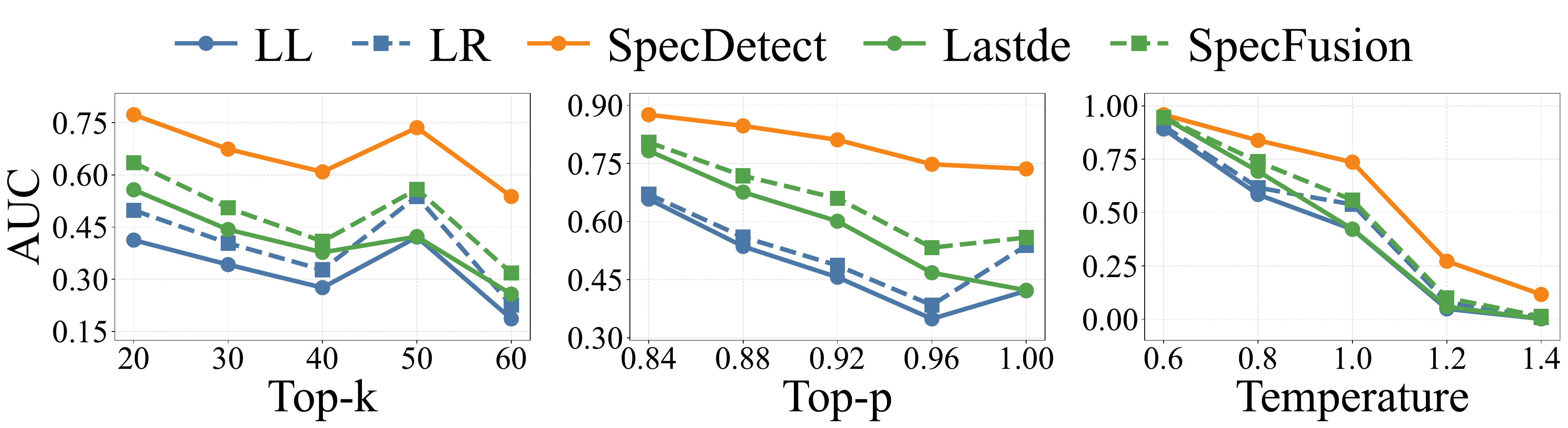}
    \caption{\textbf{WMT16 cross-language test (decoding-scope sensitivity).} Following Figure~\ref{fig:full_sensitivity}.}
    \label{fig:crosslingual_wmt16_decoding}
\end{figure}

Figure~\ref{fig:crosslingual_wmt16_length} and \ref{fig:crosslingual_wmt16_decoding} summarizes the WMT16 check. The two figures report length sensitivity and decoding-scope sensitivity for the same dataset.


\subsection{MixText Edit-Density Diagnostics}
\label{app:mixtext_density}

Edit density measures how much a modified text changes relative to its paired original. We tokenize both versions and count token-level insertions, deletions, and substitutions, with replacements counted by the larger replaced span:
\[
\mathrm{EditDensity}(x,\tilde{x}) =
\frac{\mathrm{EditOps}(x,\tilde{x})}{\max(|x|,|\tilde{x}|)} .
\]
A value near 0 indicates light local editing, while values near or above 1 indicate near-full rewriting, destructive humanizing, or substantial length expansion.

To test whether this quantity predicts detector behavior, we compare per-sample edit density with signed detector-score movement. Let $S_m(\cdot)$ denote the score of metric $m$, where larger values indicate stronger machine evidence. For pair $j$ in task $d$, we define
\[
\Delta_{j,m}^{(d)}
=
\eta_d\bigl(S_m(\tilde{x}_j)-S_m(x_j)\bigr),
\]
where $\eta_d=+1$ for Human$\to$AI operations and $\eta_d=-1$ for AI$\to$Human operations, so positive values follow the expected pairwise direction in Table~\ref{tab:mixtext_pairwise}. For each task--metric pair, we compute the Spearman correlation $\rho_{d,m}$ between edit density and $\Delta_{j,m}^{(d)}$ over the 300 paired samples, and then average $\rho_{d,m}$ across the 12 MixText tasks for each metric. The largest metric-level mean correlation is only 0.127, indicating that edit density alone weakly explains detector-score movement.

Table~\ref{tab:mixtext_density} reports the average edit density for all 12 MixText tasks, together with the best-performing main-text metric as a compact reference to Table~\ref{tab:mixtext_pairwise}.

\begin{table*}[t]
\centering
\scriptsize
\resizebox{\textwidth}{!}{%
\begin{tabular}{lllclc}
\toprule
Setting & Direction & Operation & Mean density & Best metric & Pair acc. \\
\midrule
GPT-4 & H$\to$AI & Polish-token & 0.336 & Entropy & 0.7867 \\
GPT-4 & H$\to$AI & Polish-sentence & 0.800 & Entropy & 0.7067 \\
GPT-4 & H$\to$AI & Complete & 1.089 & SpecDetect & 0.8100 \\
GPT-4 & H$\to$AI & Rewrite & 1.051 & LRR & 0.4933 \\
Llama-2 & H$\to$AI & Polish-token & 0.635 & Entropy & 0.8233 \\
Llama-2 & H$\to$AI & Polish-sentence & 0.810 & LogRank / Lastde & 0.8300 \\
Llama-2 & H$\to$AI & Complete & 0.958 & SpecDetect / Lastde & 0.9700 \\
Llama-2 & H$\to$AI & Rewrite & 0.923 & LogRank & 0.9600 \\
Human & AI$\to$H & Adapt-token & 0.098 & Entropy & 0.7733 \\
Human & AI$\to$H & Adapt-sentence & 0.228 & Lastde & 0.8467 \\
Llama-2 & AI$\to$H & Humanize & 0.985 & LogLikelihood / LogRank / Lastde / SpecFusion & 0.7333 \\
GPT-4 & AI$\to$H & Humanize & 0.981 & LogLikelihood / Lastde & 0.9667 \\
\bottomrule
\end{tabular}}
\caption{\textbf{Edit density and compact detection summary for all MixText operations.} Mean density is the average token-level edit density over 300 paired samples. The best metric and pairwise accuracy are from Table~\ref{tab:mixtext_pairwise}.}
\label{tab:mixtext_density}
\end{table*}

The table shows why edit density is informative but insufficient as a collapse threshold. Complete operations introduce sustained machine generation and favor fluctuation-based evidence, while GPT-4 rewrite has similarly high density but remains difficult. Low-density adaptation and polishing can still be detected by confidence shifts, and humanize operations reflect injected noise and edit direction rather than density alone.

\begin{figure}[t]
    \centering
    \includegraphics[width=\linewidth]{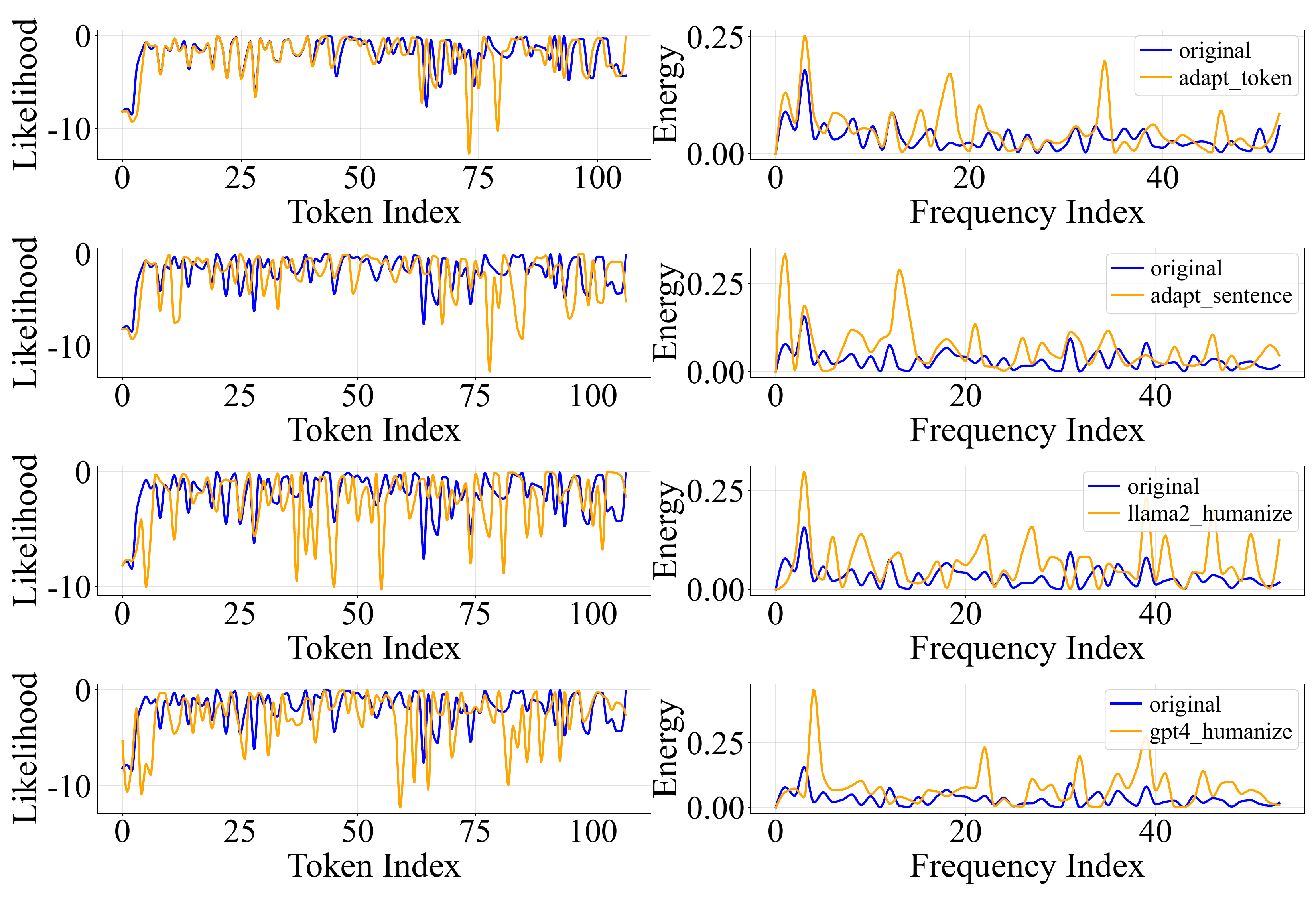}
    \caption{\textbf{Case Visualization of Signal Profiles across Humanizing Modes.} Original and humanized signals under each mode. Left: time-domain log-likelihood; right: frequency-domain spectrum.}
    \label{fig:humanize_viz}
\end{figure}

Figure~\ref{fig:humanize_viz} visualizes representative time- and frequency-domain changes for the Humanizing operations discussed in Section~\ref{sec:humanize_analysis}.

\end{document}